\pdfoutput=1

\documentclass[11pt]{article}

\usepackage[preprint]{acl}

\usepackage{times}
\usepackage{latexsym}

\usepackage[T1]{fontenc}

\usepackage[utf8]{inputenc}

\usepackage{microtype}

\usepackage{inconsolata}

\usepackage{graphicx}

\usepackage{tcolorbox}
\tcbuselibrary{skins}
\usepackage{booktabs}
\usepackage{colortbl}
\usepackage{xcolor}
\usepackage{amsmath}
\usepackage{amssymb}
\usepackage{subfigure}
\usepackage{caption}
\usepackage{float}
\usepackage{newfloat}
\usepackage{listings}
\usepackage{inconsolata}

\DeclareCaptionStyle{ruled}{labelfont=normalfont,labelsep=colon,strut=off} 
\floatstyle{ruled}
\newfloat{listing}{tb}{lst}{}
\floatname{listing}{Listing}

\definecolor{systemcolor}{HTML}{000141}
\usepackage{microtype}
\usepackage{graphicx}
\usepackage{subfigure}
\usepackage{booktabs} 
\usepackage{xcolor}

\usepackage{pgfplots}
\pgfplotsset{compat = newest}
\usepackage{multirow}
\usepackage{tcolorbox}

\usepackage{amsmath}
\usepackage{amssymb}
\usepackage{mathrsfs}
\usepackage{mathtools}
\usepackage{amsthm}
\usepackage{algorithm}
\usepackage{algpseudocode}
\usepackage{listings}
\usepackage{makecell}
\usepackage{colortbl}
\usepackage{color}
\usepackage{wrapfig}
\usepackage{cancel}
\usepackage{soul,xcolor}
\usepackage{pifont}
\usepackage{tikz}
\usepackage{tikz-cd}
\usetikzlibrary{backgrounds,circuits,circuits.ee.IEC,shapes,petri,fit,matrix}
\usepackage{adjustbox} 

\usetikzlibrary{shapes, positioning, arrows.meta, calc, decorations.pathmorphing, quotes}

\usepackage[capitalize,noabbrev]{cleveref}

\theoremstyle{plain}

\theoremstyle{definition}

\theoremstyle{remark}

\newcommand{\alink}[1]{\href{#1}{paper-link}}

\tikzstyle{simple}=[-,line width=2.000]
\tikzstyle{arrow}=[-,postaction={decorate},decoration={markings,mark=at position .5 with {\arrow{>}}},line width=1.100]
\pgfdeclarelayer{edgelayer}
\pgfdeclarelayer{nodelayer}
\pgfsetlayers{edgelayer,nodelayer,main}

\tikzstyle{none}=[inner sep=0pt]

\tikzstyle{place}=
[circle,thick,draw=blue!75,fill=blue!20,minimum size=6mm]

\tikzstyle{transition}=
[rectangle,thick,draw=black!75,fill=black!20,minimum width=1mm, minimum height=4mm]

\tikzstyle{inarrow}=[->, >=stealth, shorten >=.03cm,line width=1.5]
\tikzstyle{empty}=[circle,fill=none, draw=none]
\tikzstyle{inputdot}=[circle,fill=purple,draw=purple, scale=.25]
\tikzstyle{inputarrow}=[->,draw=purple, shorten >=.05cm]
\tikzstyle{simple}=[-,draw=purple,line width=1.000]

\makeatletter
\def\slashedarrowfill@#1#2#3#4#5{%
  $\m@th\thickmuskip0mu\medmuskip\thickmuskip\thinmuskip\thickmuskip
   \relax#5#1\mkern-7mu%
   \cleaders\hbox{$#5\mkern-2mu#2\mkern-2mu$}\hfill
   \mathclap{#3}\mathclap{#2}%
   \cleaders\hbox{$#5\mkern-2mu#2\mkern-2mu$}\hfill
   \mkern-7mu#4$%
}
\def\rightslashedarrowfill@{%
  \slashedarrowfill@\relbar\relbar\mapstochar\rightarrow}
\newcommand{\xslashedrightarrow}[2][]{%
  \ext@arrow 0055{\rightslashedarrowfill@}{#1}{#2}}

\newif\iflics

\usepackage{enumitem}
\usepackage{ amssymb }

\usepackage{listings}
\usepackage{xcolor}

\definecolor{backcolour}{rgb}{0.95, 0.95, 0.92}
\definecolor{codegreen}{rgb}{0,0.6,0}
\definecolor{codegray}{rgb}{0.5,0.5,0.5}
\definecolor{codepurple}{rgb}{0.58,0,0.82}
\definecolor{codemagenta}{rgb}{0.58,0,0.82}

\definecolor{backcolour}{rgb}{0.99, 0.99, 1.0}  
\definecolor{commentcolor}{rgb}{0.1, 0.5, 0.1}    
\definecolor{keywordcolor}{rgb}{0, 0, 0.45}       
\definecolor{stringcolor}{rgb}{0.75, 0.45, 0.45}  

\lstdefinestyle{mystyle}{
    backgroundcolor=\color{backcolour},   
    commentstyle=\color{commentcolor},
    keywordstyle=\color{keywordcolor}\bfseries,  
    stringstyle=\color{stringcolor},
    basicstyle=\tiny\ttfamily,  
    morestring=[s]{>}{<},
    morestring=[b]",
    morecomment=[s]{<?}{?>},
    breakatwhitespace=false,         
    breaklines=true,                 
    captionpos=b,                    
    keepspaces=true,                 
    showspaces=false,                
    showstringspaces=false,
    showtabs=false,                  
    tabsize=2,
    numbers=none  
}

\lstdefinelanguage{markdown}{
    morekeywords={},
    morecomment=[l]{\#},
    commentstyle=\color{codepurple},
    moredelim=[l][\textbf]{**},
    sensitive=true,
}

\definecolor{systemcolor}{HTML}{000141}
\tcbuselibrary{skins}

\definecolor{backcolour}{rgb}{0.95, 0.95, 0.92}
\definecolor{codegreen}{rgb}{0,0.6,0}
\definecolor{codegray}{rgb}{0.5,0.5,0.5}
\definecolor{codepurple}{rgb}{0.58,0,0.82}
\definecolor{codemagenta}{rgb}{0.58,0,0.82}

\definecolor{backcolour}{rgb}{0.99, 0.99, 1.0}  
\definecolor{commentcolor}{rgb}{0.1, 0.5, 0.1}    
\definecolor{keywordcolor}{rgb}{0, 0, 0.45}       
\definecolor{stringcolor}{rgb}{0.75, 0.45, 0.45}  

\lstdefinestyle{mystyle}{
    backgroundcolor=\color{backcolour},   
    commentstyle=\color{commentcolor},
    keywordstyle=\color{keywordcolor}\bfseries,  
    stringstyle=\color{stringcolor},
    basicstyle=\tiny\ttfamily,  
    morestring=[s]{>}{<},
    morestring=[b]",
    morecomment=[s]{<?}{?>},
    breakatwhitespace=false,         
    breaklines=true,                 
    captionpos=b,                    
    keepspaces=true,                 
    showspaces=false,                
    showstringspaces=false,
    showtabs=false,                  
    tabsize=2,
    numbers=none  
}

\lstdefinelanguage{markdown}{
    morekeywords={},
    morecomment=[l]{\#},
    commentstyle=\color{codepurple},
    moredelim=[l][\textbf]{**},
    sensitive=true,
}

\title{Autonomous Data Selection with Zero-shot Generative Classifiers\\ for Mathematical Texts}

\author{
 Yifan Zhang$^{1*}$, Yifan~Luo$^{1}$\thanks{Equal contribution}, \textbf{Yang Yuan}$^{1,2\dagger}$, \textbf{Andrew C Yao}$^{1,2}$\thanks{Corresponding authors} \\
 $^1$Tsinghua University,
 $^2$Shanghai Qi Zhi Institute\\
\texttt{yifanzhangresearch@gmail.com}, \texttt{luoyf24@mails.tsinghua.edu.cn}\\
\texttt{yuanyang@tsinghua.edu.cn}, \texttt{andrewcyao@tsinghua.edu.cn}
}

\begin{document}
\maketitle
\renewcommand{\thefootnote}{\fnsymbol{footnote}}
\setcounter{footnote}{3}

\begin{abstract}
We present \textit{Autonomous Data Selection} (\textbf{AutoDS}), a method that leverages base language models themselves as zero-shot ``generative classifiers'' to automatically curate high-quality mathematical texts. Unlike prior approaches that require human annotations or training a dedicated data filter, AutoDS relies solely on a model’s logits to determine whether a given passage is mathematically informative and educational. By integrating AutoDS into a continual pretraining pipeline, we substantially boost downstream performance on challenging math benchmarks (MATH, GSM8K, and BBH) while using far fewer tokens than previous methods. Empirically, our approach achieves roughly a twofold improvement in pretraining token efficiency over strong baselines, underscoring the potential of self-directed data selection in enhancing mathematical reasoning. We release our curated AutoMathText dataset to facilitate future research in automated domain-specific data curation\footnote{The code is available at \href{https://github.com/yifanzhang-pro/AutoMathText}{https://github.com/yifanzhang-pro/AutoMathText}.}. The AutoMathText dataset is available at
\href{https://huggingface.co/datasets/math-ai/AutoMathText}{https://huggingface.co/datasets/math-ai/AutoMathText}.

\end{abstract}

\section{Introduction}

Language models (LMs) have witnessed tremendous advancements, becoming increasingly adept at natural language understanding, generation, and reasoning~\citep{devlin2018bert, radford2018improving, radford2019language, brown2020language, OpenAI2023GPT4TR, anil2023palm}. Yet, integrating \emph{domain-specific} knowledge into these models remains a critical and challenging frontier~\citep{lewkowycz2022solving, azerbayev2023llemma}. Mathematical reasoning, in particular, demands specialized expertise: texts often contain symbolic formulas, multi-step derivations, and rigorous proof structures that differ considerably from conventional language tasks~\citep{hendrycks2021measuring, paster2023openwebmath, wang2023generative}. Despite the growing enthusiasm for building LMs with robust mathematical proficiency, the field continues to face a scarcity of well-curated and high-quality mathematical corpora, underscoring the urgent need for innovative approaches to create and refine domain-specific training data.

Recent efforts have begun to address this gap. For instance, \citet{gunasekar2023textbooks} and \citet{li2023textbooks} demonstrated the utility of large LMs (e.g., GPT-4) to appraise the educational value of code snippets in the Stack dataset~\citep{kocetkov2022stack}, subsequently training a traditional classifier (e.g., random forest) for data filtering. While these approaches represent a pivotal step toward more judicious data curation, they typically produce only \emph{discrete} labels (e.g., ``good'' vs.\ ``bad''), discarding the finer granularity of data quality. In mathematical contexts, subtle nuances matter: a dataset entry with an ``educational value'' of 0.95 should arguably be treated differently from one at 0.001. Relying solely on binary classification can thus limit the efficiency and precision of the training pipeline.

A promising alternative is to assign \emph{continuous} real-valued scores to each data point, thereby enabling the model to focus selectively on the most informative texts. However, constructing such a continuous scoring system poses nontrivial challenges. Large language models often struggle with generating reliable numerical values or sampling consistently from intricate distributions~\citep{hopkins2023can, hu2023amortizing}. Drawing inspiration from the Direct Preference Optimization (DPO) framework~\citep{rafailov2023direct}, we propose a simpler yet effective solution: leveraging the model’s own logits associated with targeted tokens (e.g., ``YES'' vs.\ ``NO'') to produce a quantitative score function. This approach avoids costly labeling efforts and bypasses the need for training an additional classifier on human-annotated data.

Concretely, we introduce Autonomous Data Selection (\textbf{AutoDS}), which uses zero-shot meta-prompts to evaluate the quality of mathematical texts for continual pretraining. Instead of relying on aligned or fine-tuned models, we take a strong \emph{base} model and prompt it with two yes/no questions assessing (1)~the level of “mathematical intelligence” in the text, and (2)~its utility for future math learning. From the resulting logits on “YES” and “NO,” we compute a single real-valued \textsc{LM-Score} that captures the text’s educational value. This enables a more fine-grained assessment than binary filtering approaches~\citep{li2023textbooks, paster2023openwebmath}, thus amplifying token efficiency by selectively training on the most instructive samples.

Another distinguishing factor of our method is its ability to \emph{autonomously} curate data: no separate human-annotated corpus or reward model is needed. Techniques like supervised fine-tuning (SFT)~\citep{radford2019language}, Reinforcement Learning from Human Feedback (RLHF)~\citep{ouyang2022training}, or specialized preference modeling~\citep{rafailov2023direct} are not required. By directly applying a softmax-based score on the base model’s logits, \textsc{AutoDS} orchestrates a form of active, self-directed learning, where the model itself identifies and harnesses the best materials for continual pretraining. This paves the way for a more \emph{dynamic} and \emph{scalable} data selection pipeline, especially relevant for highly specialized fields like mathematics.

\begin{figure*}[!htbp]
  \centering
  \includegraphics[width=0.8\textwidth]{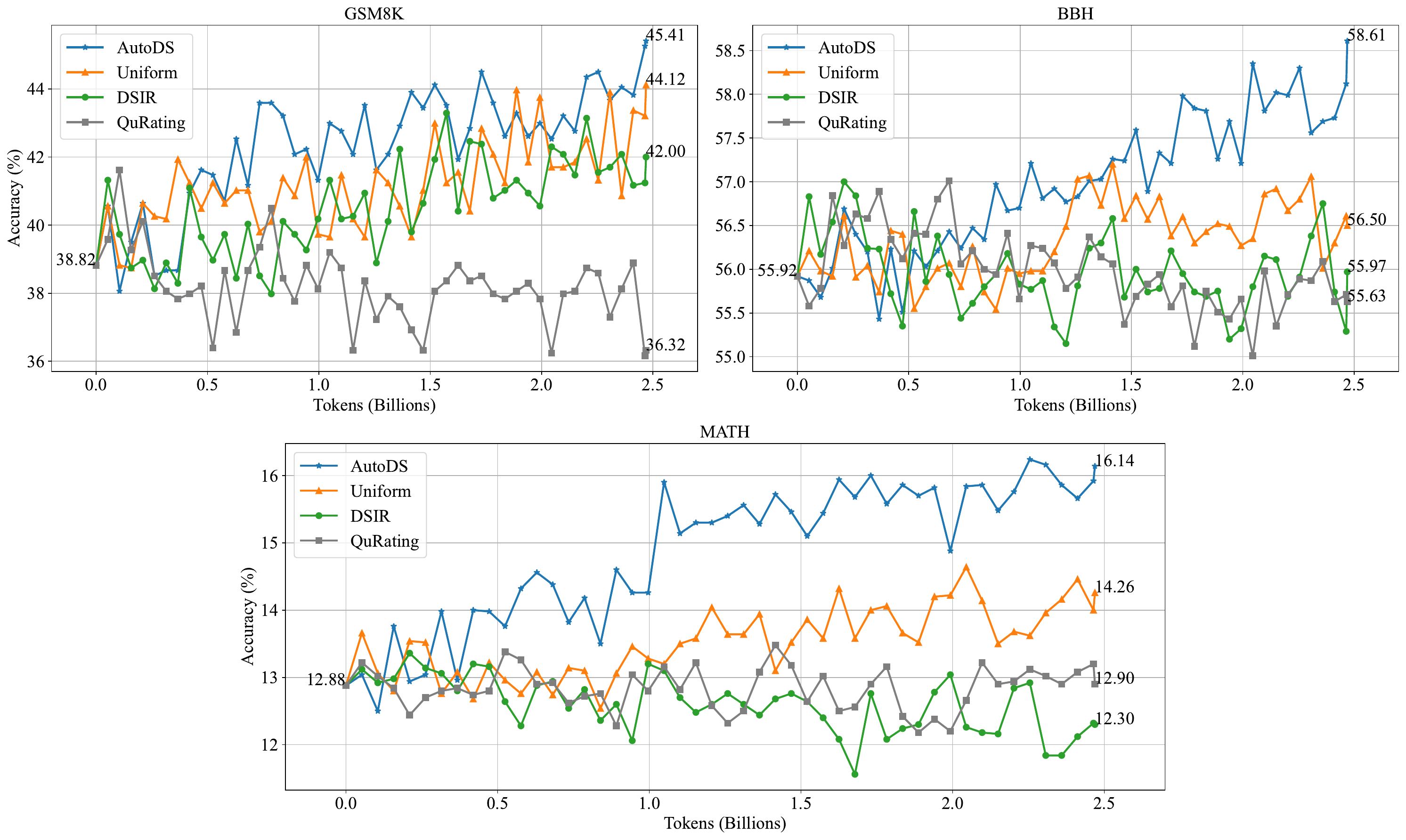}
  \caption{Visualization of Mistral-7B's performances of continual pretrained models with different data selection methods on GSM8K~\citep{hendrycks2021measuring},  BIG-Bench Hard (BBH)~\citep{suzgun2022challenging} and MATH~\citep{hendrycks2021measuring} tasks.}
  \label{fig:baseline_comparison}
\end{figure*}

Empirically, we show that continually pretraining language models on this \emph{auto-curated} dataset yields substantial gains on mathematics benchmarks, such as MATH~\citep{hendrycks2021measuring}, GSM8K~\citep{cobbe2021training}, and BIG-Bench Hard (BBH)~\citep{suzgun2022challenging}. Remarkably, these improvements come with far fewer tokens compared to previous continual pretraining works, effectively boosting training efficiency by roughly a factor of two. Figure~\ref{fig:baseline_comparison} previews these performance trends on the Mistral-7B model~\citep{jiang2023mistral}, underscoring the efficacy of our data selection method. 

Our key contributions are three-fold:
\begin{itemize}
    \item We propose a straightforward, zero-shot generative classifier framework that uses \emph{logits}-based scoring to automatically filter large-scale mathematical data. It circumvents the need for supervised or human feedback signals while retaining fine-grained control over data quality.
    \item We assemble and release a carefully curated dataset, \emph{AutoMathText}, drawn from multiple high-value sources (e.g., OpenWebMath, arXiv, Algebraic Stack). It addresses the scarcity of domain-specific mathematical corpora essential for training more powerful LMs.
    \item Through extensive evaluations, we demonstrate that LMs continually pretrained with \textsc{AutoDS} achieve significantly higher accuracy on mathematical tasks, surpassing binary-based filtering methods and achieving a 2$\times$ increase in pretraining token efficiency.
\end{itemize}

\section{Language Models as Zero-shot Generative Classifiers}
\label{sec:auto-select}

Recent advances in large language models (LLMs) have demonstrated remarkable potential for complex reasoning and decision-making~\citep{wei2022chain, bubeck2023sparks}. Building on these capabilities, we propose leveraging \emph{base} language models in a zero-shot fashion to verify whether candidate documents possess the mathematical rigor and educational utility necessary for continual pretraining. This approach goes beyond conventional few-shot paradigms~\citep{brown2020language}, which require task-specific prompt engineering or model fine-tuning, by directly harnessing the LLMs' inherent capacity to assess textual content without reliance on human annotations.

\paragraph{Generative Classifiers.}
The centerpiece of our \textsc{AutoDS} framework is a scoring function using LLMs as \emph{generative classifiers} for quantifying the model's propensity to affirm or deny the mathematical value of a given piece of text. Specifically, we examine the logits associated with ``YES'' and ``NO'' when the model is prompted with two diagnostic questions (e.g., \emph{Is this text mathematically intelligent? Is it educational for future math learning?}). Let $\mathrm{logit}(\text{YES})$ and $\mathrm{logit}(\text{NO})$ denote the output logits of the model for these tokens. We define the zero-shot LM-Score as follows:
\begin{align}
    &\operatorname{LM-Score}(\cdot) \notag\\
    = &\frac{\exp(\mathrm{logit}(\text{YES}))}
         {\exp(\mathrm{logit}(\text{YES})) + \exp(\mathrm{logit}(\text{NO}))}.
    \label{eq:lm-score}
\end{align}
Here, a higher value indicates the model's stronger inclination to judge the text as mathematically valuable. Notably, this mirrors the Bradley-Terry model from reward modeling in RLHF~\citep{ouyang2022training}, yet our method requires no supervised dataset nor explicit preference labels.

\paragraph{Zero-shot Meta-prompts.}
To elicit these logits in a consistent and interpretable manner, we formulate a concise meta-prompt~\citep{zhang2023meta} that asks two questions about each candidate text. As shown in Figure~\ref{fig:lm-score-web}, the prompt is presented in a structured format, and the model is directed to respond only with ``YES'' or ``NO.'' Crucially, we extract the logits from the underlying language model before any additional sampling. This procedure obviates the need for manual filtering or annotated corpora.

\begin{figure}[ht!]
\centering
\begin{tcolorbox}[width=0.45\textwidth,colback=blue!2!white,colframe=gray!50!blue]
    \begin{minipage}{\textwidth}
\small
{\huge ``}{\color{systemcolor} $<$system$>$

You are ChatGPT, equipped with extensive expertise in mathematics and coding, and skilled in complex reasoning and problem-solving. In the following task, I will present a text excerpt from a website. Your role is to evaluate whether this text exhibits mathematical intelligence and if it is suitable for educational purposes in mathematics. Please respond with only YES or NO 

$<$/system$>$

User: \{

\quad{}``url": ``\{url\}",

\quad{}``text": ``\{text\}"

\}

1. Does the text exhibit elements of mathematical intelligence? Respond with YES or NO

2. Is the text suitable for educational purposes for YOURSELF in the field of mathematics? Respond with YES or NO

Assistant: 1.} {\huge ''}
    \end{minipage}
\end{tcolorbox}
\caption{Illustration of our zero-shot meta-prompt designed for \textsc{AutoDS}. The underlying model is instructed to respond only with ``YES'' or ``NO,'' thereby enabling a direct extraction of logits for each answer.}
\label{fig:lm-score-web}
\end{figure}

Because the meta-prompt poses two questions, we compute the \textsc{LM-Score} by multiplying the probabilities corresponding to ``YES'' for each question:
\begin{align}
    &\operatorname{LM\-Score}(Q_1, Q_2) = \notag\\
    &\operatorname{LM-Score}(Q_1) \;\times\; \operatorname{LM-Score}(Q_2).
    \label{eq:lm-score-multiple}
\end{align}
Thus, a document must be deemed sufficiently positive on \emph{both} dimensions—mathematical intelligence and educational worth—to achieve a high overall score.

\paragraph{Autonomous Continual Pretraining.}
A critical advantage of our approach is its ease of integration into \emph{continual pretraining} pipelines. Rather than training a secondary classifier or obtaining human labels, the base model itself autonomously selects or discards documents over time. By re-evaluating each new batch of data, the model can dynamically refine its own training corpus, effectively learning ``what to learn next.'' This self-directed mechanism is especially appealing for specialized domains (e.g., mathematics), where human annotations are often scarce, expensive, or unreliable.

\paragraph{Avoiding Human Annotations.}
Finally, our zero-shot strategy obviates the necessity of extensive labeled datasets or alignment with human preferences (e.g., via RLHF). This decision reflects accumulating evidence suggesting that strong LLMs exhibit competitive (and, in many cases, superior) capacity for domain-specific judgement~\citep{burns2023weak}. This autonomy is critical for mathematics, where naive keyword-based heuristics (e.g., counting \LaTeX\ symbols) may fail to capture deeper aspects of mathematical reasoning. By leaning on the model’s emergent understanding, we thus enable more scalable, cost-efficient data curation, as demonstrated in Figure~\ref{fig:lm-score-examples-web-two-columns} and Figure~\ref{fig:lm-score-examples-code-lean-two-columns} in Appendix~\ref{sec:examples}.

\begin{figure*}[!ht]
\centering
\begin{tcolorbox}[width=0.95\textwidth,colback=cyan!2!white,colframe=gray!50!cyan]
    \begin{minipage}{0.45\textwidth}
    \color{systemcolor}
        \small
        {``}Commutative Property Of Addition. If A is an $n\times m$ matrix and O is a $m\times k$ zero-matrix, then we have: AO = O. Note that AO is the $n\times k$ zero-matrix.
        ...{''}
        
        [LM-Score ($Q_1$, $Q_2$): {\color{blue}0.946}] 

        [OWMath Classifier Score: {\color{blue} $0.767$}]
        
        \vspace{1ex}
        {``}Inequality involving sums with binomial coefficient I am trying to show upper- and lower-bounds on $\frac{1}{2^n}\sum_{i=0}^n\binom{n}{i}\min(i, n-i)$ (where $n\geq 1$) to show that it grows as $\Theta(n)$. The upper-bound is easy to get since $\min(i, n-i)\leq i$ for $i\in\{0, \dots n\}$ so that $\frac{1}{2^n}\sum_{i=0}^n\binom{n}{i}\min(i, n-i)\leq \frac{1}{2^n}\sum_{i=0}^n\binom{n}{i}i = \frac{n}{2}.$
        ...{''}
        \hspace{10ex} [LM-Score ($Q_1$, $Q_2$): {\color{blue}0.931}]

        [OWMath Classifier Score: {\color{blue} $0.999$}]

        \vspace{1ex}
        {``}The radius of convergence is half the length of the interval of convergence. We noticed that, at least in the case of the geometric series, there was an interval in which it converged, but it didn’t converge at the endpoints. Show that the following alternating harmonic series converges: Series of Both Positive and Negative Terms Theorem: Convergence of Absolute Values Implies Convergence If $\sum |a_n|$ converges, then so does $\sum a_n$. Let $f: [1, \infty) \to \mathbb{R}_{+}$ be a non-negative ...
        {''}
        \hspace{35ex} [LM-Score ($Q_1$, $Q_2$): {\color{blue}0.923}]

        [OWMath Classifier Score: {\color{blue} $0.906$}]
        \end{minipage}\hfill 
    \begin{minipage}{0.45\textwidth}
    \small
    \color{systemcolor}
        {``}\# User talk:173.79.37.192
        \#\# March 2009
        Welcome to Wikipedia. Although everyone is welcome to make constructive contributions to Wikipedia, at least one of your recent edits, such as the one you made to Reaction time, did not appear to be constructive and has been reverted. Please use the sandbox for any test edits you would like to make, and read the welcome page to learn more about contributing constructively to this encyclopedia. Thank you. Hotcrocodile (talk) 01:33, 11 March 2009 (UTC)
        If this is a shared IP address, and you didn't make any unconstructive edits, consider creating an account for yourself so you can avoid further irrelevant warnings.
        \#\# NAYLA MATTHW [1] [[Media:Example.oggfhf... {''}
        \hspace{40ex} [LM-Score ($Q_1$, $Q_2$): {\color{red} $1.58 \times 10^{-5}$}]

        [OWMath Classifier Score: {\color{blue} $0.612$}]

        \vspace{3ex}
        {``}
        I've just had one recent comment flag declined on a noisy comment.
        This comment was a reply to a deleted '+1' comment and said simply:
        @FrankL Thanks! {''}
        \hspace{35ex} [LM-Score ($Q_1$, $Q_2$): {\color{red} $1.21 \times 10^{-5}$}]

        [OWMath Classifier Score: {\color{blue} $0.830$}]
    \end{minipage}
\end{tcolorbox}
\caption{Several examples on selecting web texts. The first example in the left column is from `track-it.nz', while the second one in the left column is from `math.stackexchange.com', and the third one in the left column is from `bwni.pw'. In the right column, the first example is from `wikipedia.org', and the second one is from `math.stackexchange.com'. The trained classifier (denoted as OWMath Classifier) used in OpenWebMath~\citep{paster2023openwebmath} may mainly focus on how many latex symbols, \$ and digits exist in the text, and the examples in the right column show that it may not be very effective.}
\label{fig:lm-score-examples-web-two-columns}
\end{figure*}

In summary, our zero-shot generative classification technique exploits a model’s intrinsic capacity to rate documents for mathematical utility without any additional training. This paradigm paves the way for \emph{self-supervised} data selection, drastically reducing the need for hand-labeled resources and potentially accelerating the development of LLMs proficient in mathematical reasoning.


\section{Autonomous Data Selection with Language Models}
\label{sec:auto_data_selection}

Building on the zero-shot verification approach outlined in Section~\ref{sec:auto-select}, we apply our \textit{LM-Score}-based data selection pipeline to three principal sources of mathematical texts: 
\begin{enumerate}
    \item \textbf{OpenWebMath}~\citep{paster2023openwebmath}: A curated subset of Common Crawl, already filtered for general mathematical content;
    \item \textbf{arXiv} (from RedPajama)~\citep{together2023redpajama}: Scholarly papers encompassing diverse STEM disciplines;
    \item \textbf{Algebraic Stack}~\citep{kocetkov2022stack, azerbayev2023llemma}: A specialized subset of GitHub (the ``Stack'' dataset), featuring code and discussions related to algebraic geometry.
\end{enumerate}
These sources cover a wide range of mathematical domains and difficulty levels, making them well-suited for continual pretraining.

\paragraph{Experiment Details.}
We process a total of 11.26\,M documents, amounting to over 200\,GB of data. Following the methodology presented in Section\,3, we obtain an \emph{LM-Score} for each document using the \mbox{Qwen-72B} base language model~\citep{bai2023qwen} and retain documents scoring above specified thresholds. We employ the vLLM inference framework \citep{kwon2023efficient} on nodes with A100-80G and A800-80G GPUs. The entire filtering procedure required roughly 750 GPU hours on 4\,\mbox{A100-80G} GPUs (i.e., 3000\,GPU hours total), including both loading and inference. By contrast, expert manual annotation of 11.26\,M documents (at around \$1 per document) would cost well over \$10\,M. Using standard commercial cloud pricing for GPU compute (\$2/hour for an A100), our method’s budget remains under \$10K, drastically reducing labeling cost while avoiding the pitfalls of rule-based or purely keyword-based filtering.

\paragraph{Visualization of Data Composition.}
\label{subsec:composition}

\begin{figure*}[t!]
  \centering
    \begin{subfigure}
        \centering
        \includegraphics[width=0.425\textwidth]{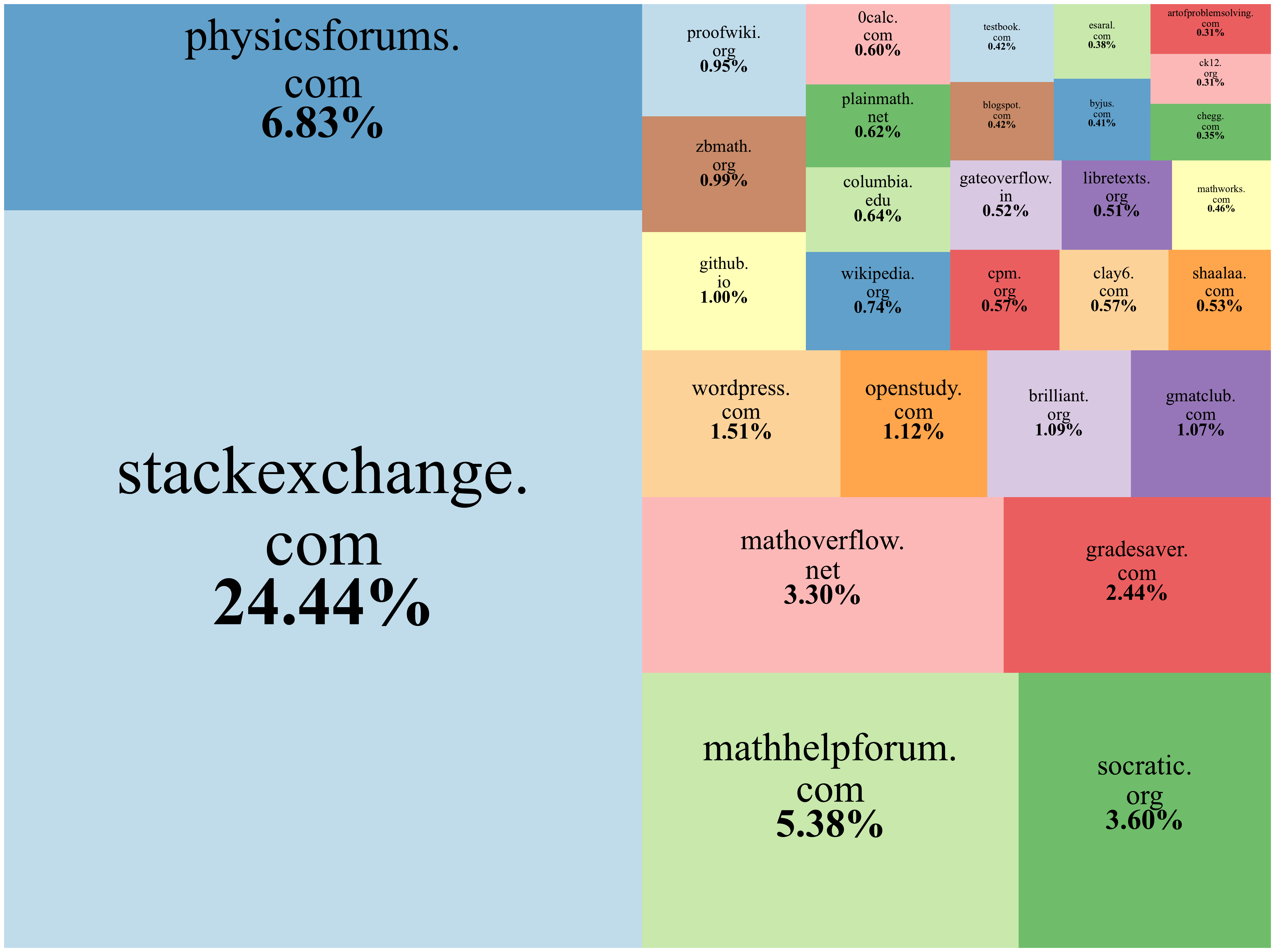}
        \label{fig:tree_map/0.50-1.00}
    \end{subfigure}
    \begin{subfigure}
        \centering
        \includegraphics[width=0.425\textwidth]{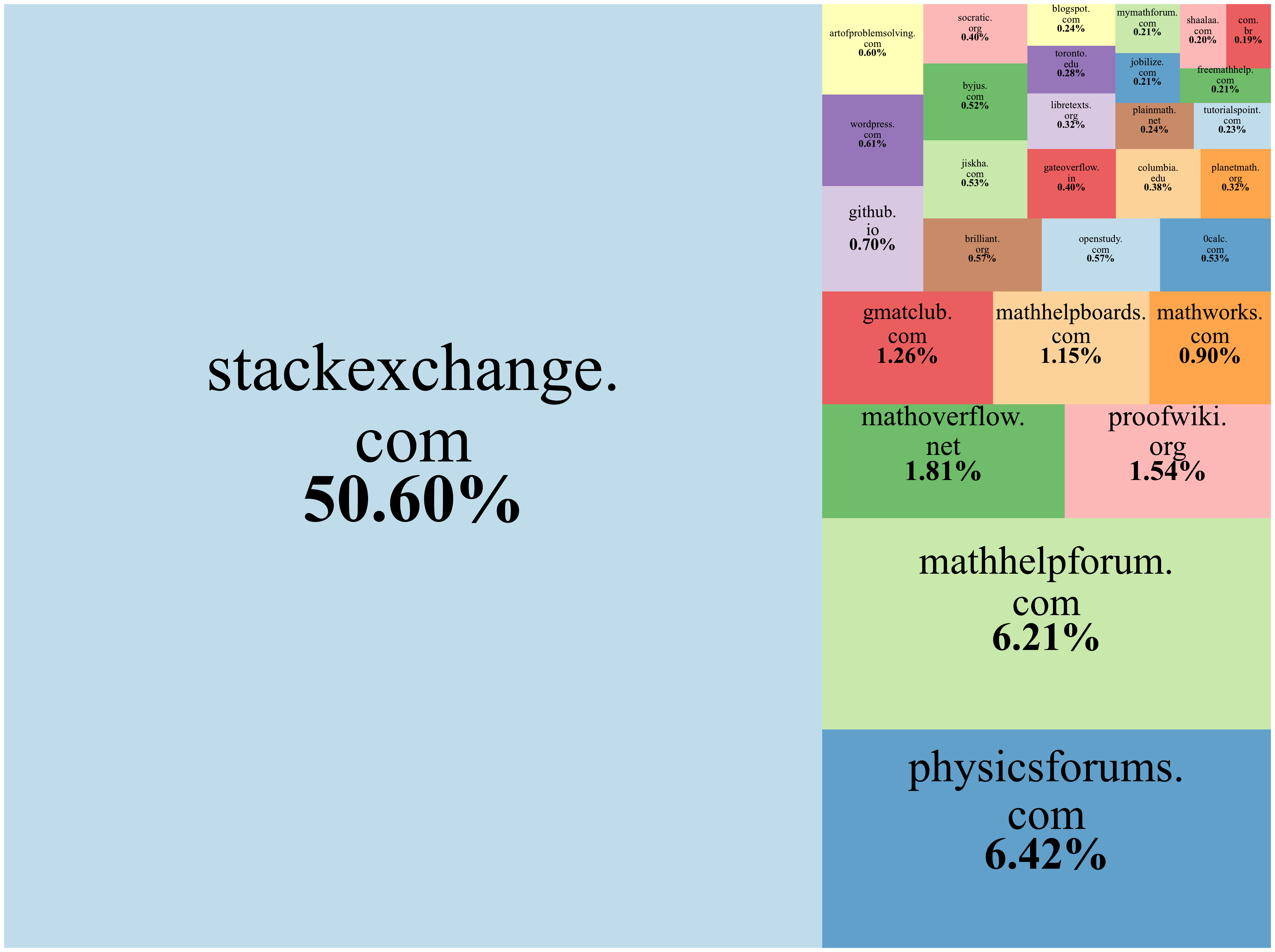}
        \label{fig:tree_map/0.75-1.00}
    \end{subfigure}
  \caption{Data composition visualization for the top-30 domains. The left treemap displays documents with LM-Scores of 0.50--1.00, while the right focuses on 0.75--1.00. StackExchange sites form a large proportion of high-score texts, many of which remain underexplored in existing math corpora.}
  \label{fig:data_composition_top30}
\end{figure*}

Examining how the selected data are distributed across different websites and content types provides insight into the quality and variety of the resulting corpus. In Figure~\ref{fig:data_composition_top30}, we plot a tree map showing the top~30 domains that scored in two different LM-Score ranges. As indicated, \texttt{*.stackexchange.com} contributes a substantial share of high-scoring examples, many of which are not yet fully leveraged in other open-source math corpora \citep{wang2023generative, liu2024augmenting}.

\begin{figure*}[t]
  \centering
  \includegraphics[width=0.825\textwidth]{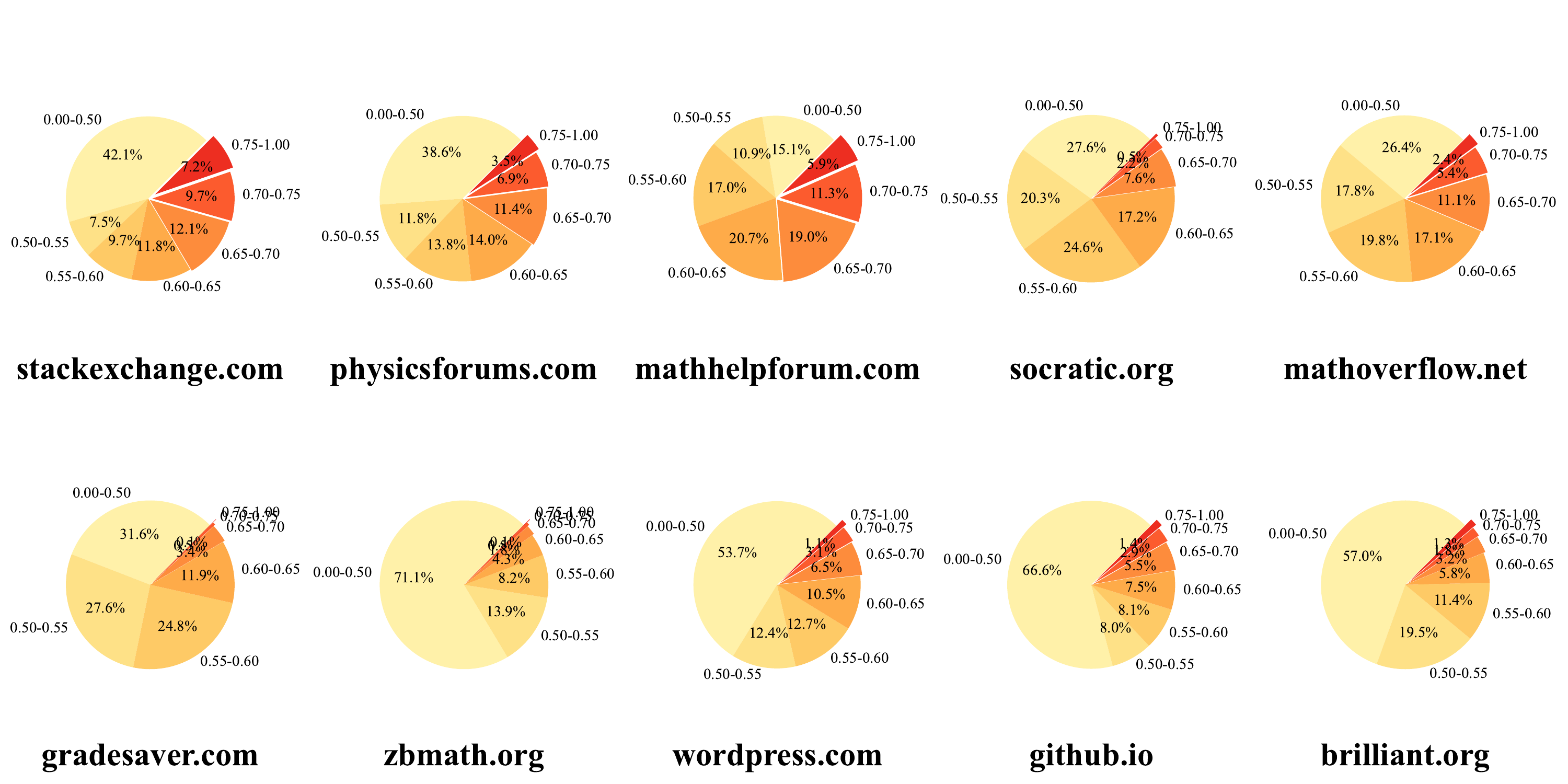}
  \caption{Distribution of LM-Scores among the top-10 domain occurrences, highlighting varying quality levels across sources.}
  \label{fig:distribution_pie}
\end{figure*}

Figure~\ref{fig:distribution_pie} offers a more granular breakdown of the highest-frequency domains and the proportion of documents falling into high-scoring bins (e.g., 0.75--1.00). We observe that many math-intensive websites such as \texttt{math.stackexchange.com} and \texttt{mathhelpforum.com} have a particularly large share of high-scoring data, underscoring their suitability for enhancing advanced mathematical language modeling.

\section{Experiments}

\definecolor{emphcolor}{rgb}{0.8902, 0.8902, 0.9804}
\definecolor{cmsensecolor}{rgb}{1, 0.800, 0.800}
\definecolor{worldknowcolor}{rgb}{1, 0.9804, 0.8039}
\definecolor{readcomcolor}{rgb}{0.8, 1, 0.8118}
\newcommand{\myemph}[1]{\cellcolor[HTML]{e3e3fa}{#1}}
\newcommand{\cmsense}[1]{\cellcolor[HTML]{ffcccc}{#1}}
\newcommand{\worldknow}[1]{\cellcolor[HTML]{fffacd}{#1}}
\newcommand{\readcom}[1]{\cellcolor[HTML]{ccffcf}{#1}}

\begin{table*}[!htb]
\small
 \caption{MATH test accuracy after continual pretraining and fine-tuning using different data (OpenWebMath and our selected data AutoMathText using method AutoDS).}
 \vspace{-1ex}
  \centering
  \begin{tabular}{c|l|c|c|c}
    \toprule
    LM-Score     & Type     & \# Tokens (M) & MATH Acc. (CPT) (\%) & MATH Acc. (SFT) (\%) \\
    \midrule
    -   & Baseline (w/o pretraining)  & 0    & \underline{12.88} & \underline{27.20}  \\
    \midrule
    - & OpenWebMath  & 328.9     & 10.50 &  26.98   \\
     0.75-1.00                       & \textbf{AutoDS}      & 328.9  & \myemph{\textbf{13.68}}  & \myemph{\textbf{28.06}}    \\
    \bottomrule
  \end{tabular}
  \label{tab:eval-math-loss}
\end{table*}

\begin{table*}[ht]
\centering
\small
\caption{Comparison of continual pretrained models using different data selection methods on complex reasoning tasks, showcasing the notable superiority of the AutoDS method.}
\label{tab:complex-reasoning}

\begin{tabular}{l|ccc}
\toprule
\textbf{Model \& Selection Method} & \textbf{MATH (5-shot)} & \textbf{GSM8K (5-shot)} & \textbf{BIG-Bench Hard (3-shot)} \\
\midrule


Gemma-2B Base & \underline{10.96} & \underline{17.29} & 34.19 \\
\quad + Uniform (OpenWebMath) & 10.16 & \myemph{\textbf{18.88}} & \myemph{\textbf{36.34}} \\
\quad + DSIR & 5.62 & 11.90 & 34.43 \\
\quad + QuRating & 9.76 & 13.19 & 31.76 \\
\quad + AutoDS & \myemph{\textbf{11.02}} & \myemph{\textbf{18.88}} & \underline{34.88} \\

\midrule


LLaMA2-7B Base & 2.94 & 12.51 & 39.89 \\
\quad + Uniform (OpenWebMath) & \underline{5.14} & \underline{19.79} & \underline{41.53} \\
\quad + DSIR & 2.56 & 12.51 & 39.49 \\
\quad + QuRating & 2.90 & 10.54 & 39.27 \\
\quad + AutoDS & \myemph{\textbf{7.74}} & \myemph{\textbf{21.99}} & \myemph{\textbf{42.76}} \\

\midrule

Mistral-7B Base & 12.88 & 38.82 & 55.92 \\
\quad + Uniform (OpenWebMath) & \underline{14.26} & \underline{44.12} & \underline{56.50} \\
\quad + DSIR & 12.30 & 42.00 & 55.97 \\
\quad + QuRating & 12.90 & 36.32 & 55.63 \\
\quad + AutoDS & \myemph{\textbf{16.14}} & \myemph{\textbf{45.41}} & \myemph{\textbf{58.61}} \\

\bottomrule
\end{tabular}
\end{table*}
In this section, we empirically assess the effectiveness of our proposed \textsc{AutoDS} method in enhancing mathematical reasoning through continual pretraining. We demonstrate that our approach substantially improves performance on several math-focused tasks while using significantly fewer tokens than previous works. We further compare \textsc{AutoDS} against existing baselines and evaluate its broader impact on general reasoning tasks.

\subsection{Experiment Details}

\paragraph{Base Models.} We consider multiple base language models: Gemma-2B~\citep{team2024gemma}, LLaMA2-7B~\citep{touvron2023llama2}, and Mistral-7B~\citep{jiang2023mistral}. These models represent mid-scale LMs frequently used in research and industry. Throughout our experiments, all methods use the same hyperparameters and training schedule for fair comparisons.

\paragraph{Datasets for Continual Pretraining.} 
We continually pretrain each model on selected portions of mathematical text. Our proposed \textsc{AutoDS} filtering (\S\ref{sec:auto-select}--\S\ref{sec:auto_data_selection}) retains only the top-scoring documents (based on LM-Score) from three major sources:
\begin{enumerate}
    \item OpenWebMath~\citep{paster2023openwebmath}, 
    \item arXiv (from RedPajama)~\citep{together2023redpajama}, 
    \item Algebraic Stack~\citep{kocetkov2022stack, azerbayev2023llemma}.
\end{enumerate}
We focus primarily on the Web subset for this work, applying zero-shot verification via Qwen-72B~\citep{bai2023qwen} to compute the LM-Scores. In total, we process over 11.26~M documents (200~GB). Documents that exceed specified LM-Score thresholds are included in the final dataset, which we call \emph{AutoMathText}.

\begin{table*}[ht]
\centering
\caption{Comprehensive comparison of continual pretrained models across diverse reasoning and comprehension tasks. The table is divided into two major sections: world knowledge and reading comprehension.}
\label{tab:other-reasoning}
\resizebox{1.0\textwidth}{!}{%
\begin{tabular}{l|ccccccc|c}
\toprule
 &  \textbf{NQ} & \multicolumn{1}{c|}{\textbf{$\text{MMLU}_{\text{STEM}}$}} & \textbf{ARC-E} & \textbf{ARC-C} & \textbf{SciQ} & \textbf{LogiQA} & \textbf{BoolQ}  &  \\ 
 \textbf{Model \& Selection Method} 
 &  \textbf{(5)} & \multicolumn{1}{c|}{\textbf{(5)}} & \textbf{(25)} & \textbf{(25)} & \textbf{(2)} & \textbf{(2)} & \textbf{(0)} & \textbf{Average} \\
\midrule




Gemma-2B Base  & 14.88 & 36.60 & 77.61 & 46.50 & 96.30 & 25.35 & 69.54 & 42.92 \\
\quad + Uniform (OpenWebMath) & 13.80 & 36.54 & 77.40 & 45.39 & 96.40 & 26.27 & 68.35 & \underline{42.95} \\
\quad + DSIR & 13.27 & 34.44 & 77.27 & 45.82 & 95.90 & 23.50 & 54.92 & 39.71 \\
\quad + QuRating & 14.32 & 33.81 & 77.95 & 46.76 & 96.20 & 24.42 & 68.53 & 41.67 \\
\quad + AutoDS & 13.27 & 36.09 & 76.81 & 46.08 & 96.10 & 27.19 & 71.28 & \myemph{\textbf{43.16}} \\

\midrule

LLaMA2-7B Base  & 26.01 & 37.08 & 80.72 & 49.74 & 96.80 & 26.57 & 77.68 & 44.99 \\
\quad + Uniform (OpenWebMath) & 26.07 & 40.09 & 80.77 & 50.09 & 96.70 & 27.65 & 78.41 & \underline{46.62} \\
\quad + DSIR & 25.76 & 36.63 & 80.98 & 48.98 & 96.50 & 26.73 & 72.54 & 44.27 \\
\quad + QuRating & 25.96 & 37.84 & 80.43 & 50.60 & 96.80 & 27.65 & 77.71 & 44.97 \\
\quad + AutoDS & 25.84 & 40.66 & 80.09 & 49.74 & 96.70 & 27.96 & 77.19 & \myemph{\textbf{47.07}} \\

\midrule

Mistral-7B Base  & 29.81 & 52.39 & 84.68 & 57.25 & 97.40 & 30.26 & 83.58 & 54.30 \\
\quad + Uniform (OpenWebMath) & 29.17 & 52.17 & 84.18 & 56.66 & 97.20 & 31.03 & 83.82 & \underline{54.91} \\
\quad + DSIR & 29.22 & 52.62 & 84.72 & 57.25 & 97.30 & 30.26 & 73.76 & 53.54 \\
\quad + QuRating & 28.89 & 52.01 & 85.48 & 57.76 & 97.30 & 31.18 & 82.81 & 54.03 \\
\quad + AutoDS & 29.06 & 52.30 & 84.18 & 55.20 & 96.80 & 31.03 & 83.12 & \myemph{\textbf{55.19}} \\
\bottomrule
\end{tabular}%
}
\end{table*}

\paragraph{Data Selection Baselines.} We compare \textsc{AutoDS} with:
\begin{enumerate}
    \item \textbf{Uniform (OpenWebMath):} Uniformly sampled data from OpenWebMath~\citep{paster2023openwebmath}, which itself was curated by simple heuristics and a trained classifier.
    \item \textbf{DSIR}~\citep{xie2023data}: A KL-divergence-based data selection approach that compares source datasets to a target domain. Here, we use the Pile’s Wikipedia split~\citep{pile} as the target to compute domain relevance.
    \item \textbf{Qurating}~\citep{wettig2024qurating}: A reward-model-based method that ranks candidate training examples by educational value, selecting those with the highest scores.
\end{enumerate}

\paragraph{Training Setup.} 
We use the codebase from {LLaMA-Factory}~\citep{llama-factory} alongside DeepSpeed ZeRO-2 Stage~\citep{deepspeed} to train on nodes with 8 A800 GPUs. The global batch size is set to 256. We conduct smaller-scale experiments with $0.3$--$0.4$B tokens for a preliminary evaluation and larger-scale experiments up to $\sim2.5$B tokens for more extensive comparisons. For Mistral-7B pretraining on $\sim 300$M tokens, we use a cosine learning rate schedule peaking at $5\times 10^{-6}$ with a 3\% warm-up ratio.
For the $2.5$B-token experiments, we use a constant learning rate of $1\times 10^{-6}$ for Mistral-7B and $1\times 10^{-5}$ for Gemma-2B, following recommended practices for continual pretraining.

We evaluate at every 100 updates (about 52M tokens) using a standard evaluation harness~\citep{eval-harness,open-llm-leaderboard-v1}.

\subsection{Continual Pretraining Results}
\label{subsec:continual-pretraining-results}

\begin{figure}[t!]
    \centering
    \includegraphics[width=0.85\columnwidth]{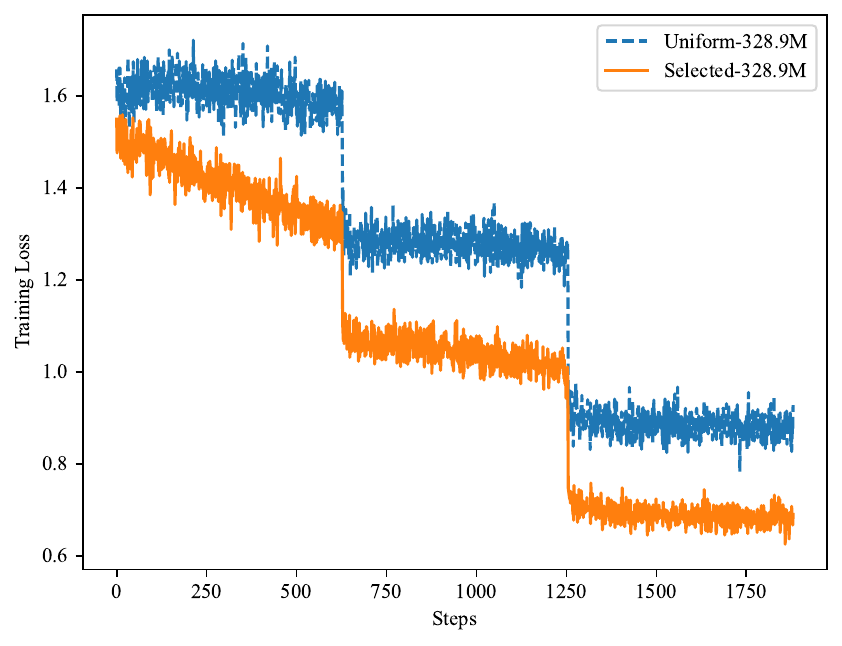}
    \caption{Training-loss evolution when continuing Mistral-7B with either uniform OpenWebMath or \textsc{AutoDS} (328.9M tokens). The loss drops more quickly for our auto-selected data.}
    \label{fig:loss_evolution}
\end{figure}

\paragraph{Preliminary Evaluation on Mistral-7B.} 
We first conduct a smaller-scale experiment on Mistral-7B-v0.1, continually pretrained with three epochs of either \emph{Uniform (OpenWebMath)} or our top-scoring subset (\emph{AutoMathText}, 328.9M tokens). Figure~\ref{fig:loss_evolution} plots the training loss evolution. We observe that \textsc{AutoDS} data yield faster and more pronounced drops in perplexity. Table~\ref{tab:eval-math-loss} reports the zero-shot MATH test accuracy both before and after supervised fine-tuning (SFT) on MetaMathQA~\citep{yu2309metamath}. The \textsc{AutoDS} filter consistently outperforms uniform sampling (e.g., $13.68\%\rightarrow16.14\%$ vs. $10.50\%\rightarrow14.26\%$ on MATH), demonstrating that higher-quality data selection facilitates stronger mathematical reasoning.

\paragraph{Larger-Scale Training (2.5B Tokens).}
Next, we scale up to $\sim2.5$B tokens of math data, comparing four methods: \textit{Uniform (OpenWebMath)}, DSIR, Qurating, and \textsc{AutoDS}. We fine-tune Gemma-2B, LLaMA2-7B, and Mistral-7B for one epoch. Figure~\ref{fig:baseline_comparison} visualizes the relative improvements on GSM8K~\citep{cobbe2021training}, BBH~\citep{suzgun2022challenging}, and MATH~\citep{hendrycks2021measuring}. Table~\ref{tab:complex-reasoning} confirms that our auto-selected data offer consistently stronger performance, particularly on MATH and GSM8K. Notably, on Mistral-7B, \textsc{AutoDS} achieves 16.14\% MATH accuracy, surpassing the uniform baseline at 14.26\% and demonstrating about 2.36$\times$ higher token efficiency. 

\subsection{Evaluation on Broader Tasks}
\label{subsec:broader-tasks}

We also investigate how improvements in mathematical reasoning might transfer to other cognitive domains such as commonsense and reading comprehension. Table~\ref{tab:other-reasoning} reports results on three representative tasks: CommonsenseQA, ARC-Challenge, and OpenBookQA. While the gains are less pronounced than in math-focused benchmarks, \textsc{AutoDS} tends to either match or slightly surpass the baselines, confirming that focusing on high-quality math data does not degrade general language capabilities.


\section{Related Work}

\noindent\textbf{Mathematical Language Models and Datasets.}
Recent advances in chain-of-thought prompting~\citep{radford2019language, wei2022chain, wang2022self, fu2022complexity, gao2023pal, yao2023tree, zhang2023meta, gou2023tora} have substantially improved the reasoning capacity of large language models (LLMs). However, most efforts in this line of research focus on eliciting latent reasoning skills through prompting alone rather than augmenting such skills by continuously pretraining on specialized corpora. The creation of high-quality mathematical datasets has played a key role in driving LLMs toward more sophisticated mathematical comprehension and problem-solving abilities. Foundational work in this domain includes the AMPS dataset~\citep{hendrycks2021measuring}, which benchmarks multi-step mathematics questions, and Proof-Pile~\citep{azerbayev2023proofnet}, which provides a large-scale corpus of mathematical texts and proofs. Building upon these resources, the Llemma model~\citep{azerbayev2023llemma} specifically targets continual pretraining on math-oriented data, including OpenWebMath~\citep{paster2023openwebmath}, to refine complex reasoning skills. Despite such progress, efficiently identifying and leveraging the most instructive mathematical data remains an ongoing challenge.

\noindent\textbf{Data Selection for Language Modeling.}
Data selection strategies have been explored extensively to improve training efficiency and effectiveness in language modeling. Early efforts by \citet{brown2020language} and \citet{chowdhery2023palm} filtered large-scale web data using binary classifiers to favor more reliable or domain-relevant content (e.g., Wikipedia and books). More targeted approaches incorporate domain-specific filtering methods or heuristics: for instance, Minerva~\citep{lewkowycz2022solving} applies rules for identifying mathematical text, while DSIR~\citep{xie2023data} employs importance sampling based on KL divergence to adapt a general corpus to a desired domain. In parallel, DoReMi~\citep{xie2023doremi} optimizes domain weights with a proxy model to reduce worst-case excess loss; however, its assumption of relatively high perplexity data may not hold for math or code corpora, whose entropy is inherently lower. Elsewhere, \citet{gunasekar2023textbooks} and \citet{li2023textbooks} used GPT-4 to annotate the educational value of code data and then trained a random forest classifier for data filtering. Qurating~\citep{wettig2024qurating} proposes a reward-model-based approach to rank training examples automatically. In contrast to these techniques, the present work introduces a fully autonomous data selection framework that relies solely on zero-shot generative classification by base language models, foregoing any reliance on human or model-labeled supervision.

\section{Conclusion}
We introduced \textbf{AutoDS}, an autonomous data selection framework that transforms base language models into zero-shot ``generative classifiers'' for filtering mathematical texts. By relying solely on model logits to assign real-valued scores, our approach avoids the need for human-annotated labels and enables more fine-grained curation than conventional binary classification methods. Through extensive evaluations, we found that continually pretraining language models on this self-selected corpus markedly enhances mathematical reasoning skills while consuming significantly fewer tokens. Moreover, the improvements on MATH, GSM8K, and BBH underscore the effectiveness of AutoDS in identifying and prioritizing instructive content. Looking ahead, we plan to extend AutoDS to broader domains to further explore how self-supervised data selection can advance specialized NLP tasks. We hope that making our \emph{AutoMathText} dataset publicly available will foster further research on scalable and autonomous data selection approaches for domain-specific training.

\section*{Limitations}
Although \textbf{AutoDS} effectively curates mathematical texts without human annotations, it depends on the reliability of a large base language model’s logits, which may introduce bias when selecting or discarding documents. Furthermore, while we observe improvements on standard math benchmarks, the framework’s performance gains may not seamlessly transfer to other specialized domains without careful prompt engineering or domain adaptation. 

\section*{Ethical Statement}
All data used in this work come from publicly accessible sources, and personal or sensitive content is excluded whenever possible. Although our approach may inherit biases from underlying models and data, caution is advised when applying it in high-stakes or real-world settings.

\bibliography{ref}

\clearpage
\appendix

\section{More Related Work}

\noindent\textbf{Data Selection in Broader Domains.}
Beyond language modeling, data selection is broadly recognized as an effective method to improve model performance across domains such as vision and domain adaptation. The Moore-Lewis approach~\citep{moore2010intelligent, axelrod2017cynical} pioneered the use of cross-entropy differentials between domain-specific and general-purpose LMs for selective data sampling. Similarly, discrepancies in feature space and n-gram distributions have guided data selection in machine translation and other tasks~\citep{jiang2007instance, liu2019reinforced, ruder2017learning}. In computer vision, curriculum learning~\citep{bengio2009curriculum} and submodular optimization~\citep{wei2015submodularity} have provided structured ways to curate datasets, while recent prioritized selection methods~\citep{coleman2019selection, mindermann2022prioritized} refine training efficiency by focusing on examples that maximize model improvements. Our proposed method draws inspiration from these broader data selection paradigms but uniquely leverages base LLMs as zero-shot generative classifiers, providing a scalable, domain-specific selection mechanism without human or trained classifier inputs.

\begin{table}[ht!]
\centering
\caption{Comparison of continual pretrained models using different data selection methods on commonsense reasoning tasks.}
\resizebox{\columnwidth}{!}{%
\label{tab:complex-reasoning-2}
\begin{tabular}{l|ccc}
\toprule
\textbf{Model \& Selection Method} & \textbf{HellaSwag (10-shot)} & \textbf{PIQA (6-shot)} & \textbf{WinoGrande (15-shot)} \\
\midrule
Gemma-2B Base      & 48.30 & 70.67 & 60.54 \\
\quad + Uniform        & 52.91 & 76.71 & 66.38 \\
\quad + DSIR      & 52.95 & 77.15 & 66.61 \\
\quad + QuRating  & 53.10 & 77.53 & 66.38 \\
\quad + AutoDS    & 52.82 & 77.42 & 66.61 \\
\midrule
LLaMA2-7B Base      & 58.88 & 79.43 & 75.85 \\
\quad + Uniform        & 58.43 & 79.54 & 75.30 \\
\quad + DSIR      & 58.38 & 78.84 & 75.37 \\
\quad + QuRating  & 58.79 & 79.00 & 74.66 \\
\quad + AutoDS    & 58.28 & 78.18 & 74.51 \\
\midrule
Mistral-7B Base     & 62.82 & 82.10 & 81.22 \\
\quad + Uniform       & 62.21 & 82.21 & 80.19 \\
\quad + DSIR     & 63.10 & 81.94 & 81.37 \\
\quad + QuRating & 62.64 & 81.99 & 80.11 \\
\quad + AutoDS  & 62.72 & 82.21 & 80.03 \\
\bottomrule
\end{tabular}
}
\end{table}

\section{More on Experiments}

\subsection{More Experimental Results}

\subsection{Prompts}

\begin{figure}[ht!]
\vspace{-2mm}
\centering
\begin{tcolorbox}[width=0.475\textwidth,colback=blue!2!white,colframe=gray!50!blue]
    \begin{minipage}{\textwidth}
\small
\color{systemcolor}
{\huge ``}$<$system$>$

You are ChatGPT, the most capable large language model equipped with extensive expertise in mathematics and coding, particularly skilled in complex reasoning and problem-solving. In the following interaction, I will provide you with a text excerpt from the arXiv website. Your task is to evaluate whether this text contains elements of mathematical intelligence and if it is suitable for educational purposes for YOURSELF in the field of mathematics. Please respond with only YES or NO 

$<$/system$>$
\vspace{1ex}

User: \{

\qquad``Title": ``\{title\}",

\qquad``Abstract": ``\{abstract\}",

\qquad``Text": ``\{text\}"
    
\}

1. Does the text contain elements of mathematical intelligence? Reply with only YES or NO

2. Is the text suitable for educational purposes for YOURSELF in the field of mathematics? Reply with only YES or NO

Assistant: 1. {\huge ''}
    \end{minipage}
\end{tcolorbox}
\caption{Prompt for selecting the papers from arXiv.org.}
\label{fig:lm-score-arxiv}
\end{figure}

\begin{figure}[ht!]
\vspace{-2mm}
\centering
\begin{tcolorbox}[width=0.475\textwidth,colback=blue!2!white,colframe=gray!50!blue]
    \begin{minipage}{\textwidth}
\small
\color{systemcolor}
{\huge ``}$<$system$>$

You are ChatGPT, the most capable large language model equipped with extensive expertise in mathematics and coding, particularly skilled in complex reasoning and problem-solving. In the following interaction, I will provide you with a code excerpt from a website. Your task is to evaluate whether this code contains elements of mathematical intelligence and if it is suitable for educational purposes for YOURSELF in the field of mathematics. Please respond with only YES or NO

$<$/system$>$
\vspace{1ex}

User: \{

\qquad``url": ``\{url\}",
    
\qquad``text": ``\{text\}"
    
\}

1. Does the code contain elements of mathematical intelligence? Reply with only YES or NO

2. Is the code suitable for educational purposes for YOURSELF in the field of mathematics? Reply with only YES or NO

Assistant: 1. {\huge ''}
    \end{minipage}
\end{tcolorbox}
\caption{Prompt for selecting code snippets from GitHub.}
\label{fig:lm-score-github}
\vspace{-1ex}
\end{figure}

\subsection{Alternative Score functions}
\label{sec:alternative-score-functions}

One can use alternative scoring functions corresponding to different partition functions, such as the formula shown below.

\begin{equation*}
\resizebox{0.95\hsize}{!}{$\begin{aligned}
    &\operatorname{LM-Score}_{\text{alternative}}(\cdot) = \\
    &\frac{\exp(\operatorname{max}(\operatorname{logit}(\text{`YES'}), \operatorname{logit}(\text{`Yes'})))}{\exp(\operatorname{max}(\operatorname{logit}(\text{`YES'}), \operatorname{logit}(\text{`Yes'}))) + \exp(\operatorname{max}(\operatorname{logit}(\text{`NO'}), \operatorname{logit}(\text{`No'})))}.
    \end{aligned}
$}
\end{equation*}



\section{Appendix for Examples}
\label{sec:examples}

\begin{figure*}[ht!]
\centering
\begin{tcolorbox}[width=0.95\textwidth,colback=cyan!2!white,colframe=gray!50!cyan]
        {\huge ``}
\small
\color{systemcolor}
Define a function called \texttt{isOdd} that takes an argument, $n \in \mathbb{N}$, and returns a proposition that asserts that $n$ is odd. The function will thus be a predicate on values of type $\mathbb{N}$. Hint: a number is odd if it's one more than an even number.
\[
\text{def isOdd} (n :\mathbb{N}) : \text{Prop} := \exists m : \text{nat}, 2 \cdot m + 1 = n
\]
To test your predicate, use ``example'' to write and prove \texttt{isOdd(15)}.
\begin{verbatim}
example : isOdd 15 :=
begin
unfold isOdd,
apply exists.intro 7,
apply rfl,
end
\end{verbatim}
Define \texttt{isSmall} : $\mathbb{N} \rightarrow \text{Prop}$, to be a predicate that is true exactly when the argument, $n$, is such that $n = 0 \lor n = 1 \lor n = 2 \lor n = 3 \lor n = 4 \lor n = 5$. (Don't try to rewrite this proposition as an inequality; just use it as is.)
\[
\text{def isSmall} (n :\mathbb{N}) : \text{Prop} := n = 0 \lor n = 1 \lor n = 2 \lor n = 3 \lor n = 4 \lor n = 5
\]
        ...{\huge ''}
        
        [LM-Score ($Q_1$, $Q_2$): {\color{blue}0.963}] 
        
{\huge ``}
Define the universes and variables for the context of our category and functor:

\[ \text{\textbf{universes}} \ v \ u \]
\[ \text{\textbf{variables}} \ \{J : \text{Type} \ v\} \ [ \text{small\_category} \ J] \ \{C : \text{Type} \ u\} \ [ \text{category}.\{v\} \ C] \ (F : J \rightarrow C) \]

\noindent Enter noncomputable theory mode and define the initial object's colimit cocone:
\begin{verbatim}
def is_initial.colimit_cocone {j : J} (hj : is_initial j)
  [has_colimit F] [\forall (a b : J) (f : a \rightarrow b),
  is_iso (F.map f)] :
  cocone F :=
{ X := F.obj j,
  \iota :=
  { app := $\lambda$ i, inv (F.map $ hj.to _),
    naturality' := begin
      intros a b f,
      dsimp,
      simp only [is_iso.eq_inv_comp, is_iso.comp_inv_eq, 
      category.comp_id],
      simp_rw ← F.map_comp,
      congr' 1,
      apply hj.hom_ext,
    end } }
\end{verbatim}
...{\huge ''}

[LM-Score ($Q_1$, $Q_2$): {\color{blue} $0.439$}]
\end{tcolorbox}
\caption{Examples contain Lean4 code. It is difficult for human beings without math expertise to judge the educational value of these examples for language models on learning mathematics.}
\label{fig:lm-score-examples-code-lean-two-columns}
\end{figure*}

\subsection{Web Subset}

\begin{figure*}[htb]
\centering
\begin{tcolorbox}[enhanced,width=0.95\textwidth,boxsep=5pt,left=5pt,right=5pt,top=3pt,bottom=3pt,colback=white!2!white,colframe=gray!50!cyan, before skip=0pt,after skip=0pt]
    \textbf{Example: }\\
    \small
    ``\# In mathematics the monomial basis of a polynomial ring is its basis (as vector space or free module over the field or ring of coefficients) that consists in the set of all monomials. The monomials form a basis because every polynomial may be uniquely written as a finite linear combination of monomials (this is an immediate consequence of the definition of a polynomial). One indeterminate The polynomial ring K[x] of the univariate polynomial over a field K is a K-vector space, which has $1,x,x^2,x^3, \ldots$ as an (infinite) basis. More generally, if K is a ring, K[x] is a free module, which has the same basis. The polynomials of degree at most d form also a vector space (or a free module in the case of a ring of coefficients), which has $1,x,x^2,\ldots$ as a basis The canonical form of a polynomial is its expression on this basis: $a_0 + a_1 x + a_2 x^2 + \ldots + a_d x^d,$ or, using the shorter sigma notation: $\sum_{i=0}^d a_ix^i.$ The monomial basis in naturally totally ordered, either by increasing degrees $1<x<x^2<\cdots,$ or by decreasing degrees $1>x>x^2>\cdots.$ Several indeterminates In the case of several indeterminates $x_1, \ldots, x_n,$ a monomial is a product $x_1^{d_1}x_2^{d_2}\cdots x_n^{d_n},$ where the $d_i$ are non-negative integers. Note that, as $x_i^0=1$, an exponent equal to zero means that the corresponding indeterminate does not appear in the monomial; in particular $1=x_1^0x_2^0\cdots x_n^0$ is a monomial. ...''\\
    \textbf{LM-Score ($Q_1$):} {\color{blue}0.987}, \quad
    \textbf{LM-Score ($Q_2$):} {\color{blue}0.662}, \quad
    \textbf{LM-Score ($Q_1$, $Q_2$):} {\color{blue}0.653}
\end{tcolorbox}
\end{figure*}

\begin{figure*}[htb]
\centering
\begin{tcolorbox}[enhanced,width=0.95\textwidth,boxsep=5pt,left=5pt,right=5pt,top=3pt,bottom=3pt,colback=white!2!white,colframe=gray!50!cyan, before skip=0pt,after skip=0pt]
    \textbf{Example: Commutative Property Of Addition}\\
    \small
    ``Commutative Property Of Addition 2. If A is an n×m matrix and O is a m×k zero-matrix, then we have: AO = O Note that AO is the n×k zero-matrix. Matrix Matrix Multiplication 11:09. We have 1. To understand the properties of transpose matrix, we will take two matrices A and B which have equal order. The identity matrix is a square matrix that has 1’s along the main diagonal and 0’s for all other entries. In a triangular matrix, the determinant is equal to the product of the diagonal elements. This matrix is often written simply as $I$, and is special in that it acts like 1 in matrix multiplication. Is the Inverse Property of Matrix Addition similar to the Inverse Property of Addition? The identity matrices (which are the square matrices whose entries are zero outside of the main diagonal and 1 on the main diagonal) are identity elements of the matrix product. Learning Objectives. In fact, this tutorial uses the Inverse Property of Addition and shows how it can be expanded to include matrices! Keywords: matrix; matrices; inverse; additive; additive inverse; opposite; Background Tutorials. ...''\\
    \textbf{LM-Score ($Q_1$):} {\color{blue}0.991}, \quad
    \textbf{LM-Score ($Q_2$):} {\color{blue}0.954}, \quad
    \textbf{LM-Score ($Q_1$, $Q_2$):} {\color{blue}0.946}
\end{tcolorbox}
\end{figure*}

\begin{figure*}[htb]
\centering
\begin{tcolorbox}[enhanced,width=0.95\textwidth,boxsep=5pt,left=5pt,right=5pt,top=3pt,bottom=3pt,colback=white!2!white,colframe=gray!50!cyan, before skip=0pt,after skip=0pt]
    \textbf{Example: Comparing the magnitudes of expressions}\\
    \small
    ``\# Comparing the magnitudes of expressions of surds I recently tackled some questions on maths-challenge / maths-aptitude papers where the task was to order various expressions made up of surds (without a calculator, obviously). I found myself wondering whether I was relying too much on knowing the numerical value of some common surds, when a more robust method was available (and would work in more difficult cases). For example, one question asked which is the largest of: (a) $\sqrt{10}$ (b) $\sqrt2+\sqrt3$ (c) $5-\sqrt3$ In this case, I relied on my knowledge that $\sqrt{10} \approx 3.16$ and $\sqrt2\approx 1.41$ and $\sqrt3 \approx 1.73$ to find (a) $\approx 3.16$, (b) $\approx ~3.14$ and (c) $\approx ~3.27$ so that the required answer is (c). But this seemed inelegant: I felt there might be some way to manipulate the surd expressions to make the ordering more explicit. I can't see what that might be, however (squaring all the expressions didn't really help).  ...''\\
    \textbf{LM-Score ($Q_1$):} {\color{blue}0.991}, \quad
    \textbf{LM-Score ($Q_2$):} {\color{blue}0.946}, \quad
    \textbf{LM-Score ($Q_1$, $Q_2$):} {\color{blue}0.937}
\end{tcolorbox}
\end{figure*}

\begin{figure*}[htb]
\centering
\begin{tcolorbox}[enhanced,width=0.95\textwidth,boxsep=5pt,left=5pt,right=5pt,top=3pt,bottom=3pt,colback=white!2!white,colframe=gray!50!cyan, before skip=0pt,after skip=0pt]
    \textbf{Example: In Calculus, function derivatives}\\
    \small
    ``\# In Calculus, how can a function have several different, yet equal, derivatives? I've been pondering this question all night as I work through some problems, and after a very thorough search, I haven't found anything completely related to my question. I guess i'm also curious how some derivatives are simplified as well, because in some cases I just can't see the breakdown. Here is an example: $f(x) = \dfrac{x^2-6x+12}{x-4}$ is the function I was differentiating. Here is what I got: $f '(x) = \dfrac{x^2-8x+12}{(x-4)^2}$ which checks using desmos graphing utility. Now, when I checked my textbook(and Symbolab) they got: $f '(x) = 1 - \dfrac{4}{(x-4)^2}$ which also checks on desmos. To me, these derivatives look nothing alike, so how can they both be the equal to the derivative of the original function? Both methods used the quotient rule, yet yield very different results. Is one of these "better" than the other? I know that it is easier to find critical numbers with a more simplified derivative, but IMO the derivative I found seems easier to set equal to zero than the derivative found in my book.I also wasn't able to figure out how the second derivative was simplified, so I stuck with mine. I'm obviously new to Calculus and i'm trying to understand the nuances of derivatives. ...''\\
    \textbf{LM-Score ($Q_1$):} {\color{blue}0.985}, \quad
    \textbf{LM-Score ($Q_2$):} {\color{blue}0.950}, \quad
    \textbf{LM-Score ($Q_1$, $Q_2$):} {\color{blue}0.936}
\end{tcolorbox}
\end{figure*}

\begin{figure*}[htb]
\centering
\begin{tcolorbox}[enhanced,width=0.95\textwidth,boxsep=5pt,left=5pt,right=5pt,top=3pt,bottom=3pt,colback=white!2!white,colframe=gray!50!cyan, before skip=0pt,after skip=0pt]
    \textbf{Example: Math help on cubics}\\
    \small
    ``\# Math Help - working backwards - cubics 1. \#\# working backwards - cubics Write an equation that has the following roots: 2, -1, 5 Answer key: $x^3 - 6x^2 + 3x + 10 = 0$ For quadratic equations, I use the sum and product of roots, this is a cubic equation, how do I solve this? Thanks. 2. Originally Posted by shenton Write an equation that has the following roots: 2, -1, 5 Answer key: $x^3 - 6x^2 + 3x + 10 = 0$ For quadratic equations, I use the sum and product of roots, this is a cubic equation, how do I solve this? Thanks. $(x - 2)(x + 1)(x - 5)$ 3. Thanks! That turns out to be not as difficult as imagined. I thought I needed to use sum and products of roots to write the equation, it does makes me wonder a bit why or when I need to use sum and products of roots. 4. Write an equation that has the following roots: 2, -1, 5 Is there any other way to solve this other than the (x-2)(x+1)(x-5) method? If we have these roots: 1, $1 + \sqrt{2}$, $1 - \sqrt{2}$ the (x - 1) ($x - 1 - \sqrt{2}$) ($x - 1 + \sqrt{2}$) method seems a bit lenghty. When we expand (x - 1) ($x - 1 - \sqrt{2}$) ($x - 1 + \sqrt{2}$) the first 2 factors, it becomes: ($x^2 -x -x\sqrt{2} -x +1 + \sqrt{2}$) ($x -1 + \sqrt{2}$) collect like terms: ($x^2 -2x -x \sqrt{2} +1 + \sqrt{2}$) ($x -1 + \sqrt{2}$) To further expand this will be lenghty, my gut feel is that mathematicians do not want to do this - it is time consuming and prone to error. There must be a way to write an equation other than the above method. Is there a method to write an equation with 3 given roots (other than the above method)? ...''\\
    \textbf{LM-Score ($Q_1$):} {\color{blue}0.991}, \quad
    \textbf{LM-Score ($Q_2$):} {\color{blue}0.943}, \quad
    \textbf{LM-Score ($Q_1$, $Q_2$):} {\color{blue}0.935}
\end{tcolorbox}
\end{figure*}

\begin{figure*}[htb]
\centering
\begin{tcolorbox}[enhanced,width=0.95\textwidth,boxsep=5pt,left=5pt,right=5pt,top=3pt,bottom=3pt,colback=white!2!white,colframe=gray!50!cyan, before skip=0pt,after skip=0pt]
    \textbf{Example: Work and time}\\
    \small
    ``\# Work and time, when work is split into parts I'm stuck on a particular type of work and time problems. For example, 1) A,B,C can complete a work separately in 24,36 and 48 days. They started working together but C left after 4 days of start and A left 3 days before completion of the work. In how many days will the work be completed? A simpler version of the same type of problem is as follows: 2) A can do a piece of work in 14 days while B can do it in 21 days. They begin working together but 3 days before the completion of the work, A leaves off. The total number of days to complete the work is? My attempt at problem 2: A's 1 day work=1/14 and B's 1 day work= 1/21 Assume that it takes 'd' days to complete the entire work when both A and B are working together. Then, (1/14 + 1/21)*d= 1 -> d=42/5 days. But it is stated that 3 days before the completion of the work, A left. Therefore, work done by both in (d-3) days is: (1/14 + 1/21)*(42/5 - 3)= 9/14 Remaining work= 1- 9/14 = 5/14 which is to be done by B alone. Hence the time taken by B to do (5/14) of the work is: (5/14)*21 = 7.5 days. Total time taken to complete the work = (d-3) + 7.5 = 12.9 days. However, this answer does not concur with the one that is provided. My Understanding of problem 1: Problem 1 is an extended version of problem 2. But since i think i'm doing problem 2 wrong, following the same method on problem 1 will also result in a wrong answer. Where did i go wrong? ...''\\
    \textbf{LM-Score ($Q_1$):} {\color{blue}0.991}, \quad
    \textbf{LM-Score ($Q_2$):} {\color{blue}0.941}, \quad
    \textbf{LM-Score ($Q_1$, $Q_2$):} {\color{blue}0.932}
\end{tcolorbox}
\end{figure*}

\begin{figure*}[htb]
\centering
\begin{tcolorbox}[enhanced,width=0.95\textwidth,boxsep=5pt,left=5pt,right=5pt,top=3pt,bottom=3pt,colback=white!2!white,colframe=gray!50!cyan, before skip=0pt,after skip=0pt]
    \textbf{Example: Inequality Involving Sums}\\
    \small
    ``Inequality involving sums with binomial coefficient I am trying to show upper- and lower-bounds on $\frac{1}{2^n}\sum_{i=0}^n\binom{n}{i}\min(i, n-i)$ (where $n\geq 1$) in order to show that it basically grows as $\Theta(n)$. The upper-bound is easy to get since $\min(i, n-i)\leq i$ for $i\in\{0, \dots n\}$ so that $\frac{1}{2^n}\sum_{i=0}^n\binom{n}{i}\min(i, n-i)\leq \frac{1}{2^n}\sum_{i=0}^n\binom{n}{i}i = \frac{n}{2}$. Thanks to Desmos, I managed to find a lower bound, but I am struggling to actually prove it. Indeed, I can see that the function $f(n)=\frac{n-1}{3}$ does provide a lower-bound. One can in fact rewrite $\frac{n-1}{3}=\frac{1}{2^n}\sum_{i=0}^n\binom{n}{i}\frac{2i-1}{3}.$ I was thus hoping to show that for each term we have $\frac{2i-1}{3}\leq \min(i, n-i)$, but this is only true if $i\leq \frac{3n+1}{5}$ and not generally for $i\leq n$. I imagine there is a clever trick to use at some point but for some reason, I am stuck here. Any help would be appreciated, thank you! EDIT: Thank you everyone for all the great and diverse answers! I flagged River Li's answer as the "accepted" one because of its simplicity due to the use of Cauchy-Schwartz inequality, which does not require a further use of Stirling's approximation. ...''\\
    \textbf{LM-Score ($Q_1$):} {\color{blue}0.988}, \quad
    \textbf{LM-Score ($Q_2$):} {\color{blue}0.941}, \quad
    \textbf{LM-Score ($Q_1$, $Q_2$):} {\color{blue}0.931}
\end{tcolorbox}
\end{figure*}

\begin{figure*}[htb]
\centering
\begin{tcolorbox}[enhanced,width=0.95\textwidth,boxsep=5pt,left=5pt,right=5pt,top=3pt,bottom=3pt,colback=white!2!white,colframe=gray!50!cyan, before skip=0pt,after skip=0pt]
    \textbf{Example: Algebraic Manipulation}\\
    \small
    ``\# Algebraic Manipulation \#\# Definition Algebraic manipulation involves rearranging variables to make an algebraic expression better suit your needs. During this rearrangement, the value of the expression does not change. \#\# Technique Algebraic expressions aren't always given in their most convenient forms. This is where algebraic manipulation comes in. For example: \#\#\# What value of $x$ satisfies $5x+8 = -2x +43$ We can rearrange this equation for $x$ by putting the terms with $x$ on one side and the constant terms on the other. $$\begin{aligned} 5x+8 &= -2x +43 \\ 5x -(-2x) &= 43 -8 \\ 7x &= 35 \\ x &= \frac{35}{7} \\ x &= 5 \quad_\square \end{aligned}$$ Algebraic manipulation is also used to simplify complicated-looking expressions by factoring and using identities. Let's walk through an example: \#\#\# $\frac{x^3+y^3}{x^2-y^2} - \frac{x^2+y^2}{x-y}.$ It's possible to solve for $x$ and $y$ and plug those values into this expression, but the algebra would be very messy. Instead, we can rearrange the problem by using the factoring formula identities for $x^3+y^3$ and $x^2-y^2$ and then simplifying. $$\begin{aligned} \frac{x^3+y^3}{x^2-y^2} - \frac{x^2+y^2}{x-y} &= \frac{(x+y)(x^2-xy+y^2)}{(x-y)(x+y)} - \frac{x^2+y^2}{x-y} \\ &= \frac{x^2-xy+y^2 -(x^2+y^2)}{x-y} \\ &= \frac{-xy}{x-y} \end{aligned}$$ Plugging in the values for $xy$ and $x-y$ gives us the answer of $3$. ...''\\
    \textbf{LM-Score ($Q_1$):} {\color{blue}0.990}, \quad
    \textbf{LM-Score ($Q_2$):} {\color{blue}0.940}, \quad
    \textbf{LM-Score ($Q_1$, $Q_2$):} {\color{blue}0.931}
\end{tcolorbox}
\end{figure*}

\begin{figure*}[htb]
\centering
\begin{tcolorbox}[enhanced,width=0.95\textwidth,boxsep=5pt,left=5pt,right=5pt,top=3pt,bottom=3pt,colback=white!2!white,colframe=gray!50!cyan, before skip=0pt,after skip=0pt]
    \textbf{Example: Finding the minimum number}\\
    \small
    ``\# Finding the minimum number of students There are $p$ committees in a class (where $p \ge 5$), each consisting of $q$ members (where $q \ge 6$).No two committees are allowed to have more than 1 student in common. What is the minimum and maximum number of students possible? It is easy to see that the maximum number of student is $pq$,however I am not sure how to find the minimum number of students.Any ideas? $1) \quad pq - \binom{q}{2}$ $2) \quad pq - \binom{p}{2}$ $3) \quad (p-1)(q-1)$ - Something is missing. Is every student supposed to be on a committee? –  JavaMan Aug 31 '11 at 16:24 @DJC:Not mentioned in the question,I guess we may have to consider that to get a solution. –  Quixotic Aug 31 '11 at 16:28 @DJC: For the minimum number of students this does not matter. –  TMM Aug 31 '11 at 16:30 @Thijs Laarhoven:Yes you are right but as the problem also asked for maximum number I have considered it in my solution. –  Quixotic Aug 31 '11 at 16:31 @Thijs, FoolForMath, I guess my question is, should the minimum answer be in terms of $p$ and $q$? –  JavaMan Aug 31 '11 at 16:31 For $1\leq i\leq p$, let $C_i$ be the set of students on the $i$th committee. Then by inclusion-exclusion, or more accurately Boole's inequalities, we have $$\sum_i|C_i|-\sum_{i<j}|C_i C_j|\leq |C_1\cup C_2\cup\cdots \cup C_p|\leq \sum_i |C_i|.$$ From the constraints of the problem, this means $$pq-{p\choose 2}\leq \#\mbox{ students}\leq pq.$$ - What is $j$ here?and I can't relate this with your answer.  $j$ is also a generic index that runs from $1$ to $p$. The inequalities are also known as Bonferroni inequalities (planetmath.org/encyclopedia/BonferroniInequalities.html), and can apply to cardinalities instead of probabilities. –  Byron Schmuland Sep 1 '11 at 14:10 I think the following theorem might be relevant: Theorem. Let $\mathcal{F}$ be a family of subsets of $\{1, \dots , n \}$ with the property that $|A \cap B| = 1$ for all $A,B \in \mathcal{F}$. Then $|\mathcal{F}| \leq n$. Also this theorem could be relevant as well. - For the case in which $p \le q+1$ an arrangement that yields the minimum number of students can be described as follows. Let $P = \{\langle m,n \rangle:1 \le m \le p, 1 \le n \le q+1\}$, and let $S = \{\langle m,n \rangle \in P:m < n\}$. If $P$ is thought of as a $p \times (q+1)$ grid, ...''\\
    \textbf{LM-Score ($Q_1$):} {\color{blue}0.985}, \quad
    \textbf{LM-Score ($Q_2$):} {\color{blue}0.863}, \quad
    \textbf{LM-Score ($Q_1$, $Q_2$):} {\color{blue}0.850}
\end{tcolorbox}
\end{figure*}

\begin{figure*}[htb]
\centering
\begin{tcolorbox}[enhanced,width=0.95\textwidth,boxsep=5pt,left=5pt,right=5pt,top=3pt,bottom=3pt,colback=white!2!white,colframe=gray!50!cyan, before skip=0pt,after skip=0pt]
    \textbf{Example: Applied Linear Algebra}\\
    \small
    ``Let $w_1 = (0,1,1)$. Expand \{$w_1$\} to a basis of $R^3$. I am reading the book, Applied Linear Algebra and Matrix Analysis. When I was doing the exercise of Section3.5 Exercise 7, I was puzzled at some of it. Here is the problem description: Let $w_1 = (0,1,1)$. Expand \{$w_1$\} to a basis of $R^3$. I don't understand its description well. I think it wants to get a span set like \{$(0,1,1)$, $(1,0,0)$, $(0,0,1)$\} which is a basis of $R^3$. And I check the reference answer, which is as followings: $(0,1,1)$, $(1,0,0)$, $(0,1,0)$ is one choice among many. I think what I have done is what question wants. So can anyone tell me am I right or wrong? Thanks sincerely. • I think you are right Apr 16, 2019 at 6:02 There is a kind of 'procedure' for dealing with questions of this kind, namely to consider the spanning set $\left\{ w_1, e_1, e_2, e_3\right\}$. Consider each vector from left to right. If one of these vectors is in the span of the previous one/s, then throw it out. If not, keep it. So in this case, we start by keeping $w_1$. Moving to the next vector, $e_1$ is not in the span of $w_1$, so we keep it as well. Moving to the next, $e_2$ is not in the span of the previous two vectors so we keep it as well. Now, considering the vector $e_3$ we see that it is in fact in the span of the previous three vectors, since $e_3 = w_1 - e_2.$ So we throw out the vector $e_3$ and end up with the basis $\left\{ w_1, e_1, e_2\right\}$. This explains the solution in the reference answer. Your solution is also correct, however. $\begin{bmatrix} 1 & 0 & 0 \\ 0 & 1 & 1 \\ 0 & 0 & 1\end{bmatrix}$ has independent rows. Hence you have found $3$ independent vectors in $\mathbb{R}^3$, that is it spans $\mathbb{R}^3$ and it forms a basis. You are correct. {$(0,1,1),(1,0,0),(0,0,1)$} is a basis of $\mathbb R^3$. Any element $(a,b,c)$ in $\mathbb R^3$ can be expressed as $a(1,0,0)+b(0,1,1)+(c-b)(0,0,1).$ If your basis is $w_1, w_2, w_3$, the textbook's choice is $w_1, w_2, w_1-w_3$ ...''\\
    \textbf{LM-Score ($Q_1$):} {\color{blue}0.964}, \quad
    \textbf{LM-Score ($Q_2$):} {\color{blue}0.882}, \quad
    \textbf{LM-Score ($Q_1$, $Q_2$):} {\color{blue}0.850}
\end{tcolorbox}
\end{figure*}

\begin{figure*}[htb]
\centering
\begin{tcolorbox}[enhanced,width=0.95\textwidth,boxsep=5pt,left=5pt,right=5pt,top=3pt,bottom=3pt,colback=white!2!white,colframe=gray!50!cyan, before skip=0pt,after skip=0pt]
    \textbf{Example: Mathematical Analysis}\\
    \small
    ``\#\# Solution to Principles of Mathematical Analysis Chapter 7 Part A \#\#\# Chapter 7 Sequences and Series of Functions \#\#\#\# Exercise 1 (By analambanomenos) Let $\{f_n\}$ be a uniformly convergent sequence of bounded functions on a set $E$. For each $n$, there is a number $M_n$ such that $\big|f_n(x)\big|<M_n$ for all $x\in E$. By Theorem 7.8, there is an integer $N$ such that $\big|f_n(x)-f_N(x)\big|<1$ if $n\ge N$ for all $x\in E$. Let $$M=\max\{M_1,\ldots,M_N\}.$$ Then for $n\le N$ and $x\in E$ we have $\big|f_n(x)\big|<M+1$, and for $n\ge N$ and $x\in E$ we have $$\big|f_n(x)\big|\le\big|f_n(x)-f_N(x)\big|+\big|f_N(x)\big|<M+1.$$ That is, the $f_n$ are uniformly bounded by $M+1$ in $E$. ...''\\
    \textbf{LM-Score ($Q_1$):} {\color{blue}0.976}, \quad
    \textbf{LM-Score ($Q_2$):} {\color{blue}0.820}, \quad
    \textbf{LM-Score ($Q_1$, $Q_2$):} {\color{blue}0.800}
\end{tcolorbox}
\end{figure*}

\begin{figure*}[htb]
\centering
\begin{tcolorbox}[enhanced,width=0.95\textwidth,boxsep=5pt,left=5pt,right=5pt,top=3pt,bottom=3pt,colback=white!2!white,colframe=gray!50!cyan, before skip=0pt,after skip=0pt]
    \textbf{Example: Vector equations}\\
    \small
    ``\# Vector equations, possible to solve for x? \#\#\#\# Jonsson Hello there, In scalar algebra, I find solving for variables a useful tool. Say ohms law, I want to find $R$ so: $$U=RI \iff R = \frac{U}{I}$$ Can I do something analogous in vector equations? I.e. May I solve for $\vec{\omega}$ in equations using cross or dot products? $$\vec{v} = \vec{\omega} \times \vec{r} \iff \vec{\omega} = ?$$ or: $$\vec{\alpha} \cdot \vec{\beta} = \gamma \iff \vec{\beta} = ?$$ It would be fantastic if I could solve for vectors in some way. Hope you are able to help. Kind regards, Marius \#\#\#\# maajdl Gold Member Solving v=wxr makes sense, since this can be seen as solving 3 equations with 3 unknowns (each components). You can find the solution easily by "multiplying" both sides by r: rxv = rx(wxr) = w (r.r) - r (w.r). ...''\\
    \textbf{LM-Score ($Q_1$):} {\color{blue}0.950}, \quad
    \textbf{LM-Score ($Q_2$):} {\color{blue}0.842}, \quad
    \textbf{LM-Score ($Q_1$, $Q_2$):} {\color{blue}0.800}
\end{tcolorbox}
\end{figure*}

\begin{figure*}[htb]
\centering
\begin{tcolorbox}[enhanced,width=0.95\textwidth,boxsep=5pt,left=5pt,right=5pt,top=3pt,bottom=3pt,colback=white!2!white,colframe=gray!50!cyan, before skip=0pt,after skip=0pt]
    \textbf{Example: Linear programming}\\
    \small
    ``\# If then Constraint 2 Hello all: I want to implement the following constraint in my linear programing model: If A=B then C=1 Else C=0 I have been looking around and there are similar problems but nobody has been helpful to address the 'non equal to' condition. Thank you in advance. asked 27 Sep '14, 17:45 Chicago 33 5 accept rate: 0\% 3 As I understand the question, you want $c$ to be binary, and $c=1$ if and only if $A=B$. I will make a couple of assumptions: There is a (large) positive $M$ such that $|A-B|\le M$ for every feasible $(A,B)$. There is a (small) positive $\epsilon$ such that whenever $A\neq B$, we can assume there is a solution satisfying $|A-B| \ge \epsilon$. Here's the formulation: $$\begin{aligned} A &\le B + My - \epsilon z \\ B &\le A + Mz - \epsilon y \\ c+y+z&=1 \\ c,y,z &\in \{0,1\} \end{aligned}$$ Now, if $c=1$, then $y=z=0$. In this case, the constraints reduce to $A \le B$ and $B \le A$, so $A=B$. Otherwise, $c=0$. Then $y+z=1$. There are two cases. ...''\\
    \textbf{LM-Score ($Q_1$):} {\color{blue}0.950}, \quad
    \textbf{LM-Score ($Q_2$):} {\color{blue}0.842}, \quad
    \textbf{LM-Score ($Q_1$, $Q_2$):} {\color{blue}0.800}
\end{tcolorbox}
\end{figure*}

\begin{figure*}[htb]
\centering
\begin{tcolorbox}[enhanced,width=0.95\textwidth,boxsep=5pt,left=5pt,right=5pt,top=3pt,bottom=3pt,colback=white!2!white,colframe=gray!50!cyan, before skip=0pt,after skip=0pt]
    \textbf{Example: Distance formula}\\
    \small
    ``The distance formula is a formula that is used to find the distance between two points. These points can be in any dimension. The x-z plane is vertical and shaded pink … If observation i in X or observation j in Y contains NaN values, the function pdist2 returns NaN for the pairwise distance between i and j.Therefore, D1(1,1), D1(1,2), and D1(1,3) are NaN values.. Contents. Print the the distance between two points on the surface of earth: ----- Input the latitude of coordinate 1: 25 Input the longitude of coordinate 1: 35 Input the latitude of coordinate 2: 35.5 Input the longitude of coordinate 2: 25.5 The distance between those points is: 1480.08 Flowchart: C++ Code Editor: Contribute your code and comments through Disqus. Interactive Distance Formula applet. Distance Formula Calculator. Find the square root of that sum: $\sqrt{90} = 9.49$. In a 3 dimensional plane, the distance between points (X 1, Y 1, Z 1) and (X 2, Y 2, Z 2) are given. The distance between two points on the three dimensions of the xyz-plane can be calculated using the distance formula The distance formula is derived from the Pythagorean theorem. and: Line passing through two points. Parameters first Iterator pointing to the initial element. Distance between 2 points in 3D space calculator uses Distance between 2 points=$\sqrt{(x2-x1)^2+(y2-y1)^2+(z2-z1)^2}$ to calculate the Distance between 2 points, ...''\\
    \textbf{LM-Score ($Q_1$):} {\color{blue}0.950}, \quad
    \textbf{LM-Score ($Q_2$):} {\color{blue}0.737}, \quad
    \textbf{LM-Score ($Q_1$, $Q_2$):} {\color{blue}0.700}
\end{tcolorbox}
\end{figure*}

\begin{figure*}[htb]
\centering
\begin{tcolorbox}[enhanced,width=0.95\textwidth,boxsep=5pt,left=5pt,right=5pt,top=3pt,bottom=3pt,colback=white!2!white,colframe=gray!50!cyan, before skip=0pt,after skip=0pt]
    \textbf{Example: Estimate from below of the sine}\\
    \small
    ``\# Estimate from below of the sine (and from above of cosine) I'm trying to do the following exercise with no success. I'm asked to prove that $\sin(x) \ge x-\frac{x^3}{2}\,, \qquad \forall x\in [0,1]$ By using Taylor's expansion, it's basically immediate that one has the better estimate $\sin(x) \ge x-\frac{x^3}{6}\,, \qquad \forall x\in [0,1]$ as the tail converges absolutely, and one can check that the difference of consecutive terms is positive. I suppose then, there is a more elementary way to get the first one. Question is: how? Relatedly, the same exercise asks me to prove that $\cos(x) \le \frac{1}{\sqrt{1+x^2}}\,,\qquad \forall x\in [0,1]$ which again I can prove by using differentiation techniques. But these haven't been explained at that point of the text, so I wonder how to do it "elementary". I showed by comparison of areas that for first quadrant angles $\sin\theta\cos\theta\le\theta\le\tan\theta$ If one multiplies the left of these inequalities by $2$ it becomes $\sin2\theta<2\theta$ so we arrive at $\sin\theta\le\theta\le\tan\theta$ Rearrange the right of these inequalities to $\frac{\sin\theta}{\theta}\ge\cos\theta$ or $1-\frac{\sin\theta}{\theta}\le1-\cos\theta=2\sin^2\frac{\theta}2\le2\left(\frac{\theta}2\right)^2=\frac{\theta^2}2$ Where we have used the left of the above inequalities above. This rearranges to $\sin\theta\ge\theta-\frac{\theta^3}2$ for first quadrant angles. ...''\\
    \textbf{LM-Score ($Q_1$):} {\color{blue}0.950}, \quad
    \textbf{LM-Score ($Q_2$):} {\color{blue}0.737}, \quad
    \textbf{LM-Score ($Q_1$, $Q_2$):} {\color{blue}0.700}
\end{tcolorbox}
\end{figure*}

\begin{figure*}[htb]
\centering
\begin{tcolorbox}[enhanced,width=0.95\textwidth,boxsep=5pt,left=5pt,right=5pt,top=3pt,bottom=3pt,colback=white!2!white,colframe=gray!50!cyan, before skip=0pt,after skip=0pt]
    \textbf{Example: Force on side of pool from water}\\
    \small
    ``Force on side of pool from water Given a pool with dimensions $\ell \times w \times h \, ,$ I am trying to derive an equation that will yield the force by the water on the sides of the pool, namely $\ell\times h \quad \mathrm{or} \quad w \times h \, .$ For the side of the pool with dimensions $\ell \times h$, I started by using the familiar equation for pressure $F = PA \, .$ Plugging in the expression for hydrostatic pressure for $P$ gives $F = \rho ghA =\rho gh(\ell \times h) = \boxed{\rho g \ell h^2} \, .$ Is my reasoning, and corresponding solution correct? Hydrostatic pressure changes with height. You have just multiplied by area, which means that you have assumed it to be constant. Instead, you should integrate over the area. You'll get an extra 1/2 term for the force. – Goobs Sep 15 '15 at 4:21 As @Goobs says, the pressure force is $0$ at the top of the water line and increases to $\rho~g~y~dA$ on a surface of area $dA$ at depth $y$. Since this pressure increases linearly from $0$ to $\rho~g~y$ the average force on the wall is the average of the start and end: so, it is half of this value, and the total pressure is $\frac 12 \rho g h (h \ell).$ Would this be correct? $\int dF = \int_0^H\rho g A \,\,dh = \rho g\ell\int_0^H h\,\,dh = \boxed{\frac{1}{2}\rho g H^2}$ – rgarci0959 Sep 15 '15 at 4:51 Yes. For bonus points you would write it as $\int dA~\rho~g~h$ to start with, as that's one of those forces that you "know" is correct ...''\\
    \textbf{LM-Score ($Q_1$):} {\color{blue}0.987}, \quad
    \textbf{LM-Score ($Q_2$):} {\color{blue}0.662}, \quad
    \textbf{LM-Score ($Q_1$, $Q_2$):} {\color{blue}0.653}
\end{tcolorbox}
\end{figure*}

\subsection{Code Subset}

\begin{figure*}[htb]
\centering
\begin{tcolorbox}[enhanced,width=0.95\textwidth,boxsep=5pt,left=5pt,right=5pt,top=3pt,bottom=3pt,colback=white!2!white,colframe=gray!50!cyan, before skip=0pt,after skip=0pt]
    \textbf{Example: Lagrange's Interpolation Method}\\
    \small
    \begin{lstlisting}[language=Python]
    X = [0, 20, 40, 60, 80, 100] 
    Y = [26.0, 48.6, 61.6, 71.2, 74.8, 75.2] 
    n = len(X)-1 
    # Degree of polynomial = number of points - 1 
    print("X =", X) 
    print("Y =", Y, end='\n\n') 
    xp = float(input("Find Y for X = ")) 
    # For degree of polynomial 3, number of points n+1 = 4: 
    # L[1] = (x-x2)/(x1-x2) * (x-x3)/(x1-x3) * (x-x4)/(x1-x4) 
    # L[2] = (x-x1)/(x2-x1) * (x-x3)/(x2-x3) * (x-x4)/(x2-x4) 
    # L[3] = (x-x1)/(x3-x1) * (x-x2)/(x3-x2) * (x-x4)/(x3-x4) 
    # L[4] = (x-x1)/(x4-x1) * (x-x2)/(x4-x2) * (x-x3)/(x4-x3) 
    # L[i] *= (x-xj)/(xi-xj) where i, j = 1 to n+1 and j != i 
    # y += Y[i]*L[i] where i = 1 to n+1 
    # List index 0 to n 
    # ~~~~~~~~~~~~~~~~~~~~~~~~ Method 1: Using for loop ~~~~~~~~~~~~~~~~~~~~~~~~ 
    yp = 0 
    # Initial summation value 
    for i in range(n+1): 
        L = 1 
        # Initial product value 
        for j in range(n+1): 
            if j == i: 
                continue 
            # j == i gives ZeroDivisionError 
            L *= (xp - X[j]) / (X[i] - X[j]) yp += Y[i]*L 
    # ~~~~~~~~~~~~~~~~~~~ Method 2: Using numpy array, prod ~~~~~~~~~~~~~~~~~~~~ 
    from numpy import array, prod 
    X = array(X, float) 
    Y = array(Y, float) 
    yp = 0 
    for Xi, Yi in zip(X, Y): 
        yp += Yi * prod((xp - X[X != Xi]) / (Xi - X[X != Xi])) 
    \end{lstlisting}
    \textbf{LM-Score ($Q_1$):} {\color{blue}0.977}, \quad
    \textbf{LM-Score ($Q_2$):} {\color{blue}0.959}, \quad
    \textbf{LM-Score ($Q_1$, $Q_2$):} {\color{blue}0.937}
\end{tcolorbox}
\end{figure*}

\begin{figure*}[htb]
\centering
\begin{tcolorbox}[enhanced,width=0.95\textwidth,boxsep=5pt,left=5pt,right=5pt,top=3pt,bottom=3pt,colback=white!2!white,colframe=gray!50!cyan, before skip=0pt,after skip=0pt]
    \textbf{Example: Scientific Computing Theory}\\
    \small
    \begin{lstlisting}[language=Python]
    # Question 01, Lab 04 
    # AB Satyaprakash - 180123062 
    # imports --------------------------------------------------------------------- 
    from sympy.abc import x 
    from sympy import cos, exp, pi, evalf, simplify 
    # functions --------------------------------------------------------------------- 
    def midpointRule(f, a, b): 
        return ((b-a)*f.subs(x, (b-a)/2)).evalf() 
    
    def trapezoidalRule(f, a, b): 
        return (((b-a)/2)*(f.subs(x, a)+f.subs(x, b))).evalf() 
    
    def simpsonRule(f, a, b): 
        return (((b-a)/6)*(f.subs(x, a)+4*f.subs(x, (a+b)/2)+f.subs(x, b))).evalf() 
    
    # program body 
    # part (a) I = integrate cosx/(1+cos^2x) from 0 to pi/2 -- exact value = 0.623225 
    f = cos(x)/(1 + cos(x)**2) 
    a, b = 0, pi/2 
    print('To integrate {} from {} to {}'.format(simplify(f), a, b)) 
    print('Evaluated value of integral using Midpoint rule is', midpointRule(f, a, b)) 
    print('Evaluated value of integral using Trapezoidal rule is', trapezoidalRule(f, a, b)) 
    print('Evaluated value of integral using Simpson rule is', simpsonRule(f, a, b)) 
    print('Exact value = 0.623225\n') 
    
    # part (b) I = integrate 1/(5+4cosx) from 0 to pi -- exact value = 1.047198 
    f = 1/(5 + 4*cos(x)) 
    a, b = 0, pi 
    print('To integrate {} from {} to {}'.format(simplify(f), a, b)) 
    print('Evaluated value of integral using Midpoint rule is', midpointRule(f, a, b)) 
    print('Evaluated value of integral using Trapezoidal rule is', trapezoidalRule(f, a, b)) 
    print('Evaluated value of integral using Simpson rule is', simpsonRule(f, a, b)) 
    print('Exact value = 1.047198\n') 
    
    # part (c) I = integrate exp(-x^2) from 0 to 1 -- exact value = 0.746824 
    f = exp(-x**2) 
    a, b = 0, 1 
    \end{lstlisting}
    \textbf{LM-Score ($Q_1$):} {\color{blue}0.982}, \quad
    \textbf{LM-Score ($Q_2$):} {\color{blue}0.946}, \quad
    \textbf{LM-Score ($Q_1$, $Q_2$):} {\color{blue}0.929}
\end{tcolorbox}
\end{figure*}

\begin{figure*}[htb]
\centering
\begin{tcolorbox}[enhanced,width=0.95\textwidth,boxsep=5pt,left=5pt,right=5pt,top=3pt,bottom=3pt,colback=white!2!white,colframe=gray!50!cyan, before skip=0pt,after skip=0pt]
    \textbf{Example: Fourth Order Runge-Kutta (RK4) Method}\\
    \small
    \begin{lstlisting}[language=Python]
    from numpy import exp, linspace, empty 
    f = lambda x: exp(x-2) - 3 # Analytical Solution 
    dy = lambda x, y: y+3 # Equation to be solved, y' = y+3 
    x = 2 # Lower limit, [2 
    xn = 4 # Upper limit, 4] 
    y = -2 # Initial condition, y(2) = -2 
    h = 0.1 # Width of each division, step size 
    n = int((xn-x)/h) # Number of divisions of the domain 
    # Plot Arrays 
    xp = linspace(x, xn, n+1) 
    # Divides from x to xn into n+1 points 
    yp = empty(n+1, float) 
    yp[0] = y 
    print('x \t\ty(RK4) \t\ty(Analytical)') 
    # Header of Output 
    print('%f \t% f \t% f' % (x, y, f(x))) 
    # Initial x and y 
    for i in range(1, n+1): 
        K1 = h * dy(x,y) 
        K2 = h * dy(x + h/2, y + K1/2) 
        K3 = h * dy(x + h/2, y + K2/2) 
        K4 = h * dy(x + h, y + K3) 
        y += 1/6*(K1 + 2*K2 + 2*K3 + K4) # y(x+h) = y(x) + 1/6(K1+2K2+2K3+K4) 
        yp[i] = y 
        x += h # x for next step, 
        x = x + h 
        print('%f \t% f \t% f' % (x, y, f(x))) 
        # ~~~~~~~~~~~~~~~~~~~~~~~~~ Plotting the function ~~~~~~~~~~~~~~~~~~~~~~~~~~ 
        import matplotlib.pyplot as plt # pyplot. 
        plt.plot(xp, yp, 'ro', xp, f(xp)) # Default plot is continuous blue line 
        plt.xlabel('x') 
        plt.ylabel('y') 
        plt.legend(['RK4', 'Analytical']) 
        plt.show() 
    \end{lstlisting}
    \textbf{LM-Score ($Q_1$):} {\color{blue}0.982}, \quad
    \textbf{LM-Score ($Q_2$):} {\color{blue}0.945}, \quad
    \textbf{LM-Score ($Q_1$, $Q_2$):} {\color{blue}0.928}
\end{tcolorbox}
\end{figure*}

\begin{figure*}[htb]
\centering
\begin{tcolorbox}[enhanced,width=0.95\textwidth,boxsep=5pt,left=5pt,right=5pt,top=3pt,bottom=3pt,colback=white!2!white,colframe=gray!50!cyan, before skip=0pt,after skip=0pt]
    \textbf{Example: Real roots of the quadratic equation}\\
    \small
    \begin{lstlisting}[language=Python]
    from math import sqrt 
    from numpy.testing import assert_equal, assert_allclose
    def real_quadratic_roots(a, b, c): 
    """ 
    Find the real roots of the quadratic equation a x^2 + b x + c = 0, if they exist. 
    Parameters ---------- 
    a : float Coefficient of x^2 
    b : float Coefficient of x^1 
    c : float Coefficient of x^0 
    Returns ------- 
    roots : tuple or float or None The root(s) (two if a genuine quadratic, one if linear, None otherwise) 
    Raises ------ 
    NotImplementedError If the equation has trivial a and b coefficients, so isn't solvable. 
    """ 
        discriminant = b**2 - 4.0*a*c 
        if discriminant < 0.0: 
            return None 
        if a == 0: 
            if b == 0: 
                raise NotImplementedError("Cannot solve quadratic with both a" " and b coefficients equal to 0.") 
            else: return -c / b 
        
        x_plus = (-b + sqrt(discriminant)) / (2.0*a) 
        x_minus = (-b - sqrt(discriminant)) / (2.0*a) 
        return x_plus, x_minus 
        
    def test_no_roots(): 
    """ 
    Test that the roots of x^2 + 1 = 0 are not real. 
    """ 
    roots = None 
    assert_equal(real_quadratic_roots(1, 0, 1), roots, err_msg="Testing x^2+1=0; no real roots.") 
    \end{lstlisting}
    \textbf{LM-Score ($Q_1$):} {\color{blue}0.977}, \quad
    \textbf{LM-Score ($Q_2$):} {\color{blue}0.950}, \quad
    \textbf{LM-Score ($Q_1$, $Q_2$):} {\color{blue}0.928}
\end{tcolorbox}
\end{figure*}

\end{document}